\documentclass[11pt]{article}
\usepackage[final]{acl}
\usepackage{times}
\usepackage{latexsym}
\usepackage[T1]{fontenc}
\usepackage[utf8]{inputenc}
\usepackage{microtype}
\usepackage{inconsolata}
\usepackage{hyperref}
\usepackage{url}
\usepackage{microtype}
\usepackage{inconsolata}
\usepackage{graphicx}
\usepackage{booktabs}
\usepackage{adjustbox}
\usepackage{array}
\usepackage{multirow}
\usepackage{amsmath}
\usepackage{amssymb}
\usepackage{paralist}
\usepackage{enumitem}
\usepackage{geometry}
\usepackage{subcaption}
\usepackage{tabularx}
\usepackage{xcolor}
\usepackage{arydshln}
\usepackage{caption} 
\usepackage{tcolorbox}
\usepackage{pgfplots}
\usepackage{pgfplotstable}
\pgfplotsset{compat=1.18}
\usepackage{siunitx}

\title{\textbf{DAMASHA}: \underline{D}etecting \underline{A}I in \underline{M}ixed \underline{A}dversarial Texts via \underline{S}egmentation with \underline{H}uman-interpretable \underline{A}ttribution}

\author{
    L. D. M. S. Sai Teja\textsuperscript{1} \enspace 
    N. Siva Gopala Krishna\textsuperscript{2} \enspace 
    Ufaq Khan\textsuperscript{3} \enspace \\
    \textbf{Muhammad Haris Khan}\textsuperscript{3} \enspace
    \textbf{Atul Mishra}\textsuperscript{2}\\
    \textsuperscript{1}NIT Silchar, India \enspace
    \textsuperscript{2}BML Munjal, Haryana, India \enspace
    \textsuperscript{3}MBZUAI, Abu Dhabi, UAE \\
    \texttt{lekkalad\_ug\_22@cse.nits.ac.in, gopalkrishna.nuthakki.22cse@bmu.edu.in,} \\\texttt{ufaq.khan@mbzuai.ac.ae, muhammad.haris@mbzuai.ac.ae,}\\ \texttt{atul.mishra.17pd@bmu.edu.in} \\ [1ex]
}

\begin{document}
\maketitle
\begin{abstract}
In the age of advanced large language models (LLMs), the boundaries between human and AI-generated text are becoming increasingly blurred. We address the challenge of segmenting mixed-authorship text, that is identifying transition points in text where authorship shifts from human to AI or vice-versa, a problem with critical implications for authenticity, trust, and human oversight. We introduce a novel framework, called Info-Mask for mixed authorship detection that integrates stylometric cues, perplexity-driven signals, and structured boundary modeling to accurately segment collaborative human-AI content. To evaluate the robustness of our system against adversarial perturbations, we construct and release an adversarial benchmark dataset \textbf{M}ixed-text \textbf{A}dversarial setting for \textbf{S}egmentation (\textbf{\textit{MAS}}), designed to probe the limits of existing detectors. Beyond segmentation accuracy, we introduce \textit{Human-Interpretable Attribution} (\textit{HIA}) overlays that highlight how stylometric features inform boundary predictions, and we conduct a small-scale human study assessing their usefulness. Across multiple architectures, Info-Mask significantly improves span-level robustness under adversarial conditions, establishing new baselines while revealing remaining challenges. Our findings\footnote{Dataset: \href{https://huggingface.co/datasets/saiteja33/DAMASHA}{saiteja33/DAMASHA}, \\GitHub: \href{https://github.com/saitejalekkala33/DAMASHA}{saitejalekkala33/DAMASHA}} highlight both the promise and limitations of adversarially robust, interpretable mixed-authorship detection, with implications for trust and oversight in human-AI co-authorship.
\end{abstract}

\section{Introduction}
\label{intro}
Discriminating between human-authored and AI-generated text has become a growing concern, spanning domains such as academic integrity, misinformation, and beyond. Early works primarily focused on detecting fully AI-generated documents \cite{jawahar2020automatic, ippolito2019automatic}. More recent research, however, emphasizes the greater challenges of detecting AI content in real-world settings, particularly in the presence of mixed-authorship texts and adversarial manipulation. Several comprehensive surveys \cite{yang-etal-2024-survey, goyal2023survey, wu2025survey} have reviewed the landscape of LLM-generated text detection, highlighting approaches ranging from trained classifiers to zero-shot statistical models and watermarking techniques. \citet{chaka2023detecting} and \citet{elkhatat2023evaluating} evaluate existing tools such as GPTZero and Turnitin, reporting significant reliability issues, especially under paraphrasing or Out-Of-Domain (OOD) shifts. The most recent works by \citet{tufts2024practical} provide a rigorous empirical analysis showing that high AUROC scores often failed at robust detection performance in unseen domains, particularly under adversarial prompting. This echoes earlier findings by \cite{sadasivan2023can}, which questions the reliability of both trained and zero-shot detectors. \citet{liu2025context} proposed \emph{In‑Context Watermarking}, embedding signals via prompt engineering without model decoding access, presenting a scalable and model-agnostic solution. Altogether, these studies highlight the progress that has been made, achieving reliable and generalizable detection across diverse domains and adversarial settings remains as open challenge.

The work of \citet{jin2020bert} reveals that authorship attribution, content moderation, and text classification all rely critically on differentiating text origins to ensure reliability. Previous studies have addressed sentence segmentation on clean texts where there are no adversarial perturbations in the text \cite{read2012sentence, koshorek2018text, morris2020textattack, ebrahimi-etal-2018-hotflip}, and adversarial robustness in NLP \cite{alzantot2018generating, ribeiro2018semantically} however, limited research focused on segmenting mixed human-AI texts under adversarial conditions. \textbf{\textit{This gap highlights the need for robust segmentation methods that are resistant to syntactic manipulations.}} Syntactical adversarial attacks make the model weaker in ensuring correct accuracy, which can degrade segmentation performance \cite{li2020textshield, li-etal-2021-contextualized}. The work of \citet{garg2020bae, jia2017adversarial} reveals that such attacks pose unique challenges in mixed-text environments, where human and AI styles intertwine, thereby necessitating advanced techniques to maintain segmentation accuracy.
To address these challenges, we present a unified framework, termed Info-Mask, for segmentation in adversarially perturbed texts. Our method employs a soft-attribution mask that modulates token representations based on authorship cues, thereby allowing the model to be more transparent in its decision and robust to syntactical attacks. In addition, we introduce a large-scale benchmark to evaluate segmentation performance through rigorous experiments under adversarial conditions.

\section{Related Work}
\label{rel_work}
Research work on detecting AI-generated text has progressed along two largely separate fronts: (i) boundary detection in mixed human-AI texts, and (ii) adversarial robustness in AIGT detection as binary classification. However, to our knowledge, no prior work addresses \textbf{both} challenges simultaneously, which is adversarially robust detection of mixed-authorship segments.

\subsection{Boundary Detection in Mixed Texts}
\citet{lee2022coauthor} introduced \textit{CoAuthor}, the first benchmark for identifying segments in text jointly authored by humans and LLMs, and then after many were introduced like \textit{MixSet} by \citet{zhang2024llm}. They demonstrate that existing detectors perform poorly on mixed texts, especially against subtle revisions. Similarly, studies such as \citet{xie2023next} and \citet{zeng2024towards} explore LLM usage in storytelling and educational hybrid essays, but focus only on boundary detection without evaluating adversarial resilience. Additional works like \citet{zeng2024detecting}, \citet{kushnareva2023ai}, and \citet{wang2023seqxgpt} further analyzed boundary detection, but again without adversarial perturbation. \citet{zeng2024detecting} released a new dataset for the mixed-text detection, and also proposed a new approach \textit{TriBERT} a two step approach 1) encoder training process, 2) calculating euclidean distances for the boundary prediction between every two adjacent prototypes. \citet{kushnareva2023ai} modified the corpus that is released by \citet{dugan2020roft} (\textit{RoFT}) which contains a lot of repetitions and added new chatgpt-texts making it \textit{RoFT-chatgpt} enabling that dataset useful for the mixed-text AI detection. Lastly, \textit{the SeqXGPT} model from \citet{wang2023seqxgpt} has significant performance by extracting white-box features from the text and passing all of this to an encoder model for sequence labeling, allowing for a good text segmenting model between AI and human collaborative texts.
\subsection{Adversarial Robustness}
A growing body of literature investigates the robustness of AIGT detectors against semantic or syntactic attacks. \citet{huang2024ai} propose SCRN, a Siamese Calibrated Reconstruction Network model designed to resist character- and word-level perturbations. \citet{krishna2023paraphrasing} show that paraphrasing can drastically reduce detection accuracy and advocate for retrieval-based defenses. The framework of \citet{masrour2025damage} {\em DAMAGE} presents a data-centric augmentation strategy with active-learning, which is a significant part of their methodology, to detect syntactically humanized AI outputs. \citet{kadhim2025adversarial} further crafting embedding-based attacks targeting token-probability classifiers. Despite this progress, these works consider only pure-AIGT, without tackling segmentation in mixed-text contexts.

\subsection{Gap and Contributions}
\label{gap_contri}
Although mixed-text boundary detection and adversarially robust classification have been studied separately, no prior work addressed their intersection, which is segmenting human-AI texts while ensuring robustness against adversarial perturbations.
In contrast, our work targets \textbf{\textit{authorship-segmentation}} in fully mixed human-AI texts subjected to \textbf{\textit{syntactic adversarial attacks}}, introducing a large benchmark dataset for it. We propose a novel model with a \textbf{\textit{soft attribution masking mechanism}}, referred as an \textbf{\textit{information mask}}, that modulates token representations based on authorial attribution cues, along with new fine-grained evaluation metrics to assess adversarial segmentation. Furthermore, our framework is designed to enable \textbf{\textit{Human-Interpretable Attribution (HIA)}}, representing the first unified framework that jointly performs adversarially robust segmentation and attribution-aware detection in the mixed-authorship setting.

\section{Methodology}
\label{method}
\subsection{Mixed Text with Adversarial Attacks Dataset}\label{dataset}
Prior works on AI text detection in mixed texts have not considered adversarial attacks, with robustness studies focusing mainly on document-level binary classification between human and AI texts. We introduce \textbf{MAS}, a benchmark of \textbf{M}ixed human- and AI-authored texts with \textbf{A}dversarial attacks for fine-grained \textbf{S}egmentation of human and AI parts. MAS combines two existing mixed-text corpora without adversarial attacks \citep{zeng2024towards, wang2023m4} and extends them with new data generated by recent and more robust AI models. Specifically, \citet{zeng2024towards} provide only academic essays across six settings [HM, MH, HMH, MHM, HMHM, MHMH]\footnote{H $\rightarrow$ Human, M $\rightarrow$ Machine}, totaling 17,136 rows (train, validation, and test), while \citet{wang2023m4} contribute a single [HM] setting from multiple domains with 31,893 rows. Table~\ref{tab:priordata} provides the detailed data splits.

\begin{table}[h!]
    \centering
    \scriptsize
    \begin{tabular}{lccc}
    \toprule
    \textbf{Dataset} & \textbf{Train} & \textbf{Dev} & \textbf{Test} \\
    \midrule
    TriBERT \cite{zeng2024towards} & 12,049 & 2,527 & 2,560 \\
    M4GT \cite{wang2023m4} & 18,245 & 2,525 & 11,123 \\
    \bottomrule
    \end{tabular}
    \vspace{-5pt}
    \caption{Dataset statistics from TriBERT and M4GT with a total corpus of \textbf{49,029} rows.}
    \label{tab:priordata}
\end{table}

\vspace{-15pt}

\begin{table}[h!]
    \centering
    \resizebox{\linewidth}{!}{%
    \begin{tabular}{lccccc}
    \toprule
    \textbf{Model} & \textbf{Reddit} & \textbf{News} & \textbf{Wikipedia} & \textbf{ArXiv} & \textbf{Q\&A} \\
    \midrule
    \textit{\textbf{HM}} (GPT-4o) & 2,000 & 2,000 & 2,000 & 2,000 & 2,000 \\
    \textit{\textbf{HM}} (DeepseekV3) & 2,000 & 2,000 & 2,000 & 2,000 & 2,000 \\ 
    \midrule
    \textit{\textbf{MH}} (GPT-4o) & 2,000 & 2,000 & 2,000 & 2,000 & 2,000 \\
    \textit{\textbf{MH}} (DeepseekV3) & 957 & 1,998 & - & 2,000 & 2,000 \\ 
    \midrule
    \textit{\textbf{Mix}} (GPT-4.1-mini) & 986 & 1,000 & 981 & 998 & 971 \\
    \textit{\textbf{Mix}} (GPT-4.1) & 987 & 1,000 & 984 & 998 & 970 \\
    \bottomrule
    \end{tabular}
    }
    \vspace{-10pt}
    \caption{Our non-adversarial data distribution of generated data across domains and models having \textbf{46,830} [20,000 (HM) + 16,955 (MH) + 9,875 (Mix)] instances.}
    \label{tab:our_data}
\end{table}

\noindent We built a non-adversarial mixed-authorship dataset using human texts sourced from benchmark datasets, in three forms: Human $\rightarrow$ AI (human-written texts truncated and completed by AI), AI $\rightarrow$ Human (AI-generated prefixes followed by human continuations), and Mixed (random human sentences replaced with AI-generated ones). All texts were cleaned prior to generation, with ground-truth boundaries annotated at the word level. To ensure diversity, we sampled using \textit{GPT-4o}, \textit{GPT-4.1}, and \textit{Deepseek-V3 671B} models. Further details on human data are given in Appendix~\ref{appendix:data_creation}.

\noindent Our part of this dataset as shown in Table~\ref{tab:our_data} includes 1) human corpus from multiple domains, 2) deeply mixed texts, 3) AI text generation from open-source and closed-source AI models, comprising an overall size of non-adversarial data of \textbf{46,830} rows, and 4) finally, inclusion of several syntactical text perturbations on the complete corpus (\citet{zeng2024towards} + \citet{wang2023m4} + Ours = \textbf{95,859}) making the whole dataset size \textbf{671,013} \textit{[95,859 * 7 (1 original + 6 attacks)]}. The attacks we have taken are \textit{Misspelling, Character Substitution, Invisible Character, Punctuation substitution, Upper-Lower Swap}. More information about the attacks and why we have chosen are mentioned in the Appendix \ref{sec:adv_attacks}. For training, we split whole corpus traditionally into 70\% train, 20\% valid and 10\% test. More on data generation and the justification of attack selection are given in Appendix~\ref{appendix:data_creation}, and~\ref{sec:adv_attacks}.

\subsection{Evaluation Metrics}
For the precise evaluation in this detection, we introduce new metrics 1) \textit{\textbf{Segment-wise Boundary Detection Accuracy (SBDA)}}, and 2) \textit{\textbf{Segment Precision}}, which are designed for the partial span overlaps, as opposed to strict token-level correctness. \textbf{\textit{Span Extraction}}: Given a predicted label sequence $\hat{\mathcal{Y}} = \langle \hat{y}_1, \hat{y}_2, \ldots, \hat{y}_m \rangle$ and ground truth labels $\mathcal{Y} = \langle y_1, y_2, \ldots, y_m \rangle$, we define a \textit{span} as any maximal contiguous subsequence of tokens where the label is 1 (i.e., predicted or true AI segments). Let $\hat{S} = \{ \hat{s}_1, \ldots, \hat{s}_k \}$ be the set of predicted spans and $S = \{ s_1, \ldots, s_l \}$ the set of gold spans, where each span $s_i = (a_i, b_i)$ denotes the start and end token indices. \textbf{\textit{IoU-Based Matching}} For a predicted span $\hat{s}$ and gold span $s$, we define their token-level Intersection-over-Union (IoU) as Equation~\ref{eq:iou}. A predicted span $\hat{s}$ is considered a \textit{true positive} if there exists a gold span $s$ such that $\text{IoU}(\hat{s}, s) \geq \tau$, where $\tau$ is a predefined threshold. For the exact detection, these metrics are introduced: SBDA and segment-wise precision at multiple thresholds ($\tau \in \{0.3, 0.5, 0.7, 0.9\}$) to capture both loose and strict boundary alignment sensitivity. Using the above definitions, we compute SBDA and SegPrec as Equations~\ref{eq:sbda_sub} and \ref{eq:segpre_sub}.

\begin{equation}
    \scriptsize
    \text{IoU}(\hat{s}, s) = \frac{|\hat{s} \cap s|}{|\hat{s} \cup s|}
    \label{eq:iou}
\end{equation}

\begin{equation}
    \scriptsize
    \text{SBDA@$\tau$} = \frac{\text{\# gold spans matched (IoU $\geq$ $\tau$)}}{\text{\# total gold spans}}
    \label{eq:sbda_sub}
\end{equation}

\begin{equation}
    \scriptsize
    \text{Seg Pre@$\tau$} = \frac{\text{\# pred spans matched (IoU $\geq$ $\tau$)}}{\text{\# total pred spans}}
    \label{eq:segpre_sub}
\end{equation}

\subsection{Model Description}\label{model_desc}
\begin{figure}[h!]
    \centering
    \includegraphics[width=0.35\textwidth]{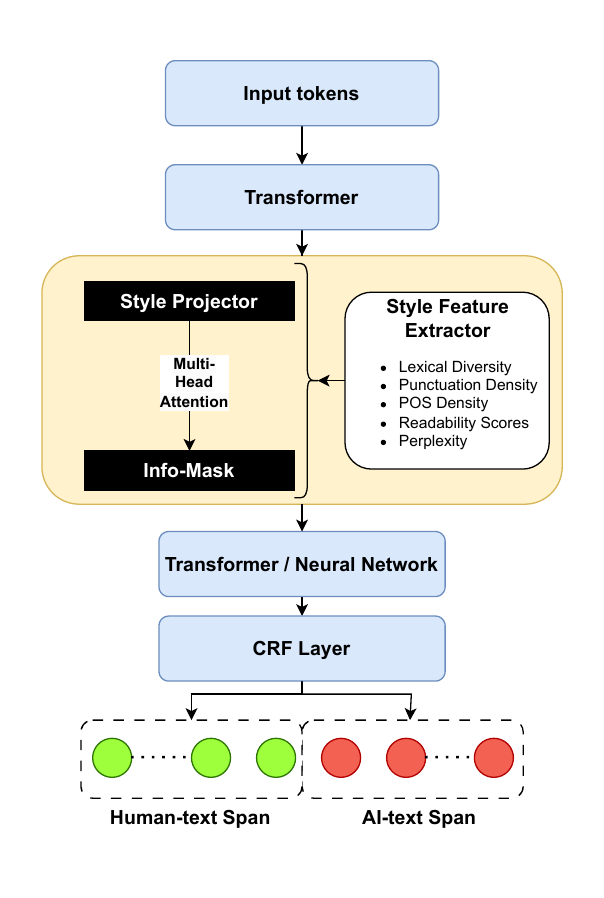}
    \caption{Model workflow showing the construction and integration of the Info-Mask to guide span segmentation using stylometric and contextual signals.}
    \label{fig:propModel}
    \vspace{-10pt}
\end{figure}

We propose a sequence segmentation framework to detect transitions between human-written and AI-generated text within a collaborative human-AI corpus. This work is especially non-trivial because the text undergoes through syntactical adversarial attacks. To detect AI-text under such attacks, our model integrates linguistically grounded attribution signals into the segmentation pipeline using a hybrid architecture combining two transformer encoders, an info-mask, and a CRF layer as shown in Figure~\ref{fig:propModel}.

\noindent \textbf{Info-Mask:} The core part of our architecture is a soft attribution masking mechanism, which dynamically modulates the influence of each token based on its syntactic and statistical alignment with typical human- or AI-generated patterns. The model computes token-level features that are rooted in structural and generative characteristics, such as \textit{perplexity}, \textit{parts-of-speech (POS tags) distributions}, \textit{Punctuation Density}, \textit{Lexical Diversity}, and \textit{Readability scores}. These linguistic features are projected and contextualized using Multi-head Attention to derive a token-wise attribution map, which serves as a soft gating mechanism over the encoder’s internal layers. This soft-gating token attribution mechanism is used to detect the presence of an adversarial attack implicitly. Unlike discrete perturbation-aware filters, this soft masking approach allows the model to selectively emphasize structurally salient regions of the text, reducing the impact of syntactic adversarial noise. Feature details and the derivation of Info-Mask are provided in Appendices~\ref{sec:info_mask_features} and \ref{appendix:info_mask_derivation}.

\noindent \textbf{CRF Tagging:} The above attribution-weighted token representations are passed through a bidirectional sequence encoder and then to a structured decoder with Conditional Random Fields (CRF), which models sequential dependencies and enforces label consistency across spans. This provides accurate segmentation of authorship boundaries, even when AI-generated segments have been adversarially crafted to resemble natural writing. By integrating syntactical attribution cues into a differentiable soft-masking and decoding architecture, the model achieves robust detection and segmentation of AI-generated content in mixed-authorship text, outperforming conventional methods in adversarially perturbed settings. The CRF loss calculation is shown in Appendix~\ref{appendix:crf_loss}.

\noindent \textbf{Human Interpretable Attribution (HIA):} During the inference, the Info-Mask also serves as an interpretability layer for human oversight. The token-level attribution maps generated can be visualized to highlight the regions that influenced the model's boundary predictions the most. This allows human users to inspect and validate predictions, correct errors, especially where adversarial attacks introduce subtle structural mimicry. The interpretability of the model is assessed by: 1) \textit{Info-mask Strength for word-to-word}, 2) \textit{Attention-x-Info-mask plot for each word}, and 3) \textit{Heat map of the extracted style features}. This transparency layer transforms the detector into an interactive system, enabling trust calibration and boundary refinement through user feedback, making it suitable for high-stakes, human-interpretable applications.

\noindent \textbf{Optimization while training:} 
As the proposed model combines two transformer backbones, a CRF layer, and the Info-Mask, training can face disruptions such as gradient explosion, diminishing updates, or uneven learning across layers. To ensure stable training and improved performance, we incorporated several optimization techniques throughout the model. They are 1) \textit{Layer Wise Learning Rate decay} implemented through the \texttt{AdamW} optimizer with grouped learning rates, 2) \textit{Dynamic Dropout} with a scheduler to avoid overfitting, 3) \textit{Gradient Clipping} to prevent gradient explosion, 4) \textit{Xavier Initialization} of the weights to maintain variance in weights across layers.

\section{Experiment Comparisons}\label{comparisions}
The proposed model has several variations as it is the combination of a transformer, neural network layer or again transformer model, and CRF layer at the end. In place of transformer and neural network positions, we can place a lot of variations. We chose three transformer models RoBERTa \cite{liu2019roberta}, ModernBERT \cite{warner2024smarter} and DeBERTa \cite{he2021debertav3} and a single neural network block Bidirectional Gated Recurrent Unit (BiGRU) for our experimentation. We experimented three different model settings 1) Transformer + CRF, 2) Transformer + NN + CRF, and 3) Transformer + Transformer + CRF, and in these model settings, the best performing model is used for the comparison with other methods or models.\\
\noindent We compared our proposed model with several baselines, including statistical-based detection methods, prior models, and zero-shot detectors, to demonstrate its reliability and improved detection metrics. Prior baselines include BiLSTM+CRF \cite{huang2015bidirectional}, SpanBERT \cite{joshi2020spanbert}, and SeqXGPT \cite{wang2023seqxgpt}. Statistical-based detection methods include \textit{logp(x)} and \textit{Entropy}, whereas zero-shot detectors include \textit{FastDetectGPT}\footnotemark[3] \cite{bao2023fast, mitchell2023detectgpt}, \textit{Glimpse}\footnotemark[3] \cite{bao2023fast}, \textit{Binoculars} \cite{hans2401spotting}, and \textit{GPTZero}\footnotemark[4].
\footnotetext[3]{\url{https://aidetect.lab.westlake.edu.cn/\#/home}}
\footnotetext[4]{\url{https://gptzero.me/}}

\begin{table*}[ht]
\centering
\scriptsize
\resizebox{1.0\linewidth}{!}{%
\begin{tabular}{llcccccccc}
\toprule
\multirow{2}{*}{\textbf{Type}} & \multirow{2}{*}{\textbf{Model}} & \multicolumn{4}{c}{\textbf{SBDA@$\tau$}} & \multicolumn{4}{c}{\textbf{Seg Pre@$\tau$}} \\
\cmidrule(lr){3-6} \cmidrule(lr){7-10}
& & \textbf{0.3} & \textbf{0.5} & \textbf{0.7} & \textbf{0.9} & \textbf{0.3} & \textbf{0.5} & \textbf{0.7} & \textbf{0.9} \\
\midrule

\multirow{2}{*}{\shortstack{Statistical\\Based}} 
& logp(x) & 0.153 & 0.107 & 0.013 & 0.005 & 0.153 & 0.107 & 0.013 & 0.005 \\
& entropy & 0.151 & 0.105 & 0.011 & 0.003 & 0.151 & 0.105 & 0.011 & 0.003 \\
\midrule

\multirow{3}{*}{\shortstack{Prior\\Baselines}}
& BiLSTM+CRF & 0.261 & 0.215 & 0.153 & 0.081 & 0.253 & 0.208 & 0.149 & 0.078 \\
& SpanBERT   & 0.294 & 0.258 & 0.190 & 0.112 & 0.281 & 0.249 & 0.183 & 0.109 \\
& SeqXGPT    & 0.315 & 0.273 & 0.228 & 0.150 & 0.301 & 0.264 & 0.219 & 0.148 \\
\midrule

\multirow{6}{*}{\shortstack{Zero-shot\\Span-Scoring}}
& GPTZero                            & 0.223 & 0.137 & 0.025 & 0.008 & 0.289 & 0.142 & 0.085 & 0.012 \\
& Binoculars                         & 0.285 & 0.235 & 0.079 & 0.022 & 0.285 & 0.235 & 0.079 & 0.022 \\
& Glimpse (baggage-002)              & 0.208 & 0.161 & 0.040 & 0.009 & 0.208 & 0.161 & 0.040 & 0.009 \\
& Glimpse (davinci-002)              & 0.214 & 0.168 & 0.044 & 0.011 & 0.214 & 0.168 & 0.044 & 0.011 \\
& FastDetectGPT (gpt-neo-2.7b)       & 0.259 & 0.208 & 0.064 & 0.015 & 0.259 & 0.208 & 0.064 & 0.015 \\
& FastDetectGPT (falcon-7b-instruct) & 0.250 & 0.214 & 0.084 & 0.019 & 0.250 & 0.214 & 0.084 & 0.019 \\
\midrule

\multirow{7}{*}{\shortstack{Proposed\\Approach}}
& RoBERTa + CRF\textsuperscript{*}            & 0.320 & 0.200 & 0.080 & 0.020 & 0.005 & 0.003 & 0.001 & 0.000 \\
& ModernBERT + CRF\textsuperscript{*}         & 0.413 & 0.407 & 0.393 & 0.358 & 0.386 & 0.380 & 0.368 & 0.335 \\
& DeBERTa + CRF\textsuperscript{*}            & 0.387 & 0.381 & 0.369 & 0.347 & 0.382 & 0.378 & 0.369 & 0.342 \\
& RoBERTa + BiGRU + CRF\textsuperscript{*}    & 0.416 & 0.408 & 0.383 & 0.352 & 0.325 & 0.319 & 0.303 & 0.275 \\
& ModernBERT + BiGRU + CRF\textsuperscript{*} & 0.419 & 0.412 & 0.398 & 0.363 & 0.402 & 0.395 & 0.382 & \textbf{0.348} \\
& DeBERTa + BiGRU + CRF\textsuperscript{*}    & 0.398 & 0.392 & 0.379 & 0.352 & 0.371 & 0.366 & 0.353 & 0.328 \\
& \textbf{RoBERTa + ModernBERT + CRF\textsuperscript{*}} & \textbf{0.457} & \textbf{0.444} & \textbf{0.421} & \textbf{0.372} & \textbf{0.414} & \textbf{0.402} & \textbf{0.387} & 0.337 \\
\bottomrule
\end{tabular}
}
\caption{Comparison of our approach with Statistical, Prior Baselines, and Zero-shot methods under SBDA and SP for different thresholds, where \textsuperscript{*} represents the model with Info-Mask Included and all our models are trained and tested on the entire corpus.}
\label{tab:sbda_sp_results}
\end{table*}

\textbf{Statistical Based Detectors.} These are based on the properties of token distributions from pretrained language models, taking them as unsupervised signals for AI-generated text detection. Specifically, we used two signals: token-level \textit{\textbf{log-likelihood}} $log p(x)$ that computes the negative log-likelihood for each token that is obtained from the encoder transformer model and token-level \textit{\textbf{entropy}} that measures the uncertainty of the model's next token-prediction. More explanation of these methods are given in Appendix \ref{appendix:stat_detect}.


\textbf{Zero-Shot Detectors.} Comparison is also done with zero-shot detectors, namely \textit{\textbf{Fast-DetectGPT}}, \textbf{\textit{Glimpse}}, \textbf{\textit{Binoculars}}, and \textbf{\textit{GPTZero}}. Except for GPTZero, these detectors were not originally designed for mixed-text sequence labeling. we adapt zero-shot detectors using a span-scoring strategy. Here, the given input text is divided/partitioned into different-length spans aligned with boundaries like sentences. Each span is evaluated holistically by the detector, and confidence scores are assigned at the span level. These scores are then back-propagated to constituent tokens, and thresholding is performed to generate binary AI–human labels. This method mitigates token-level noise through the use of coherent text segments, enabling document- and sentence-level detectors to generalize better to fine-grained boundary detection in mixed-authorship texts. Further implementation details are provided in Appendix~\ref{appendix:zeroshot_detect}.

\textbf{Prior Baselines.} \textbf{\textit{BiLSTM + CRF:}} A sequence-tagging model where a bidirectional LSTM captures contextual features and a CRF layer enforces valid tag transitions. \textbf{\textit{SpanBERT:}} An extension of BERT that masks and predicts contiguous spans, yielding stronger span-level representations for tasks like question answering. \textbf{\textit{SeqXGPT:}} A detector leveraging white-box features from LLMs, modeled as temporal sequences with convolution and self-attention, enabling both token- and sentence-level AI text detection with strong generalization.

\subsection{Proposed Model Hyperparameters}The model uses the following hyperparameters: a batch size of 64, learning rate 1e-6 $\rightarrow$ 5e-6 $\rightarrow$ 1e-5 $\rightarrow$ 1e-4 over the layer from starting to ending with a decay of 0.95, trained for 5 epochs, weight decay of 0.01, gradient clipping at 1.0, initial dropout of 0.1, dynamic dropout ranging from 0.1 to 0.3, style hidden size of 64, 5 attention heads, lexical window of 5, warm-up percentage of 0.1, cosine annealing strategy, patience of 2, 2 labels, style feature dimension of 5, and maximum sequence length of 512, and all again given in Table~\ref{tab:hyperparameters}.

\section{Results Analysis}
\label{res_analysis}

\begin{figure}[ht]
\centering
\includegraphics[width=1.0\linewidth]{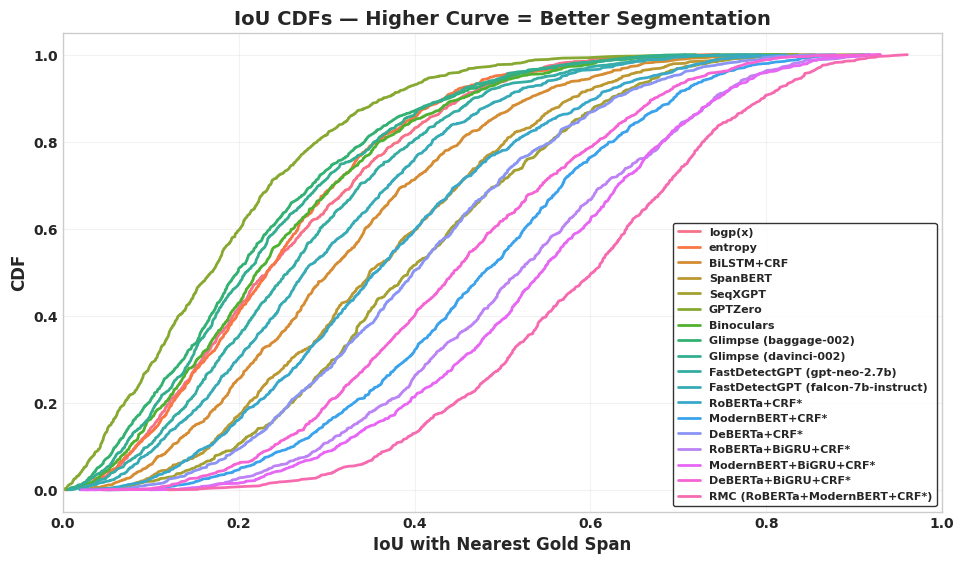}
\caption{Cumulative Distribution Function (CDF) of IoU scores for all models.}
\label{fig:iou_cdf}
\end{figure}

\begin{figure}[ht]
\centering
\includegraphics[width=1.0\linewidth]{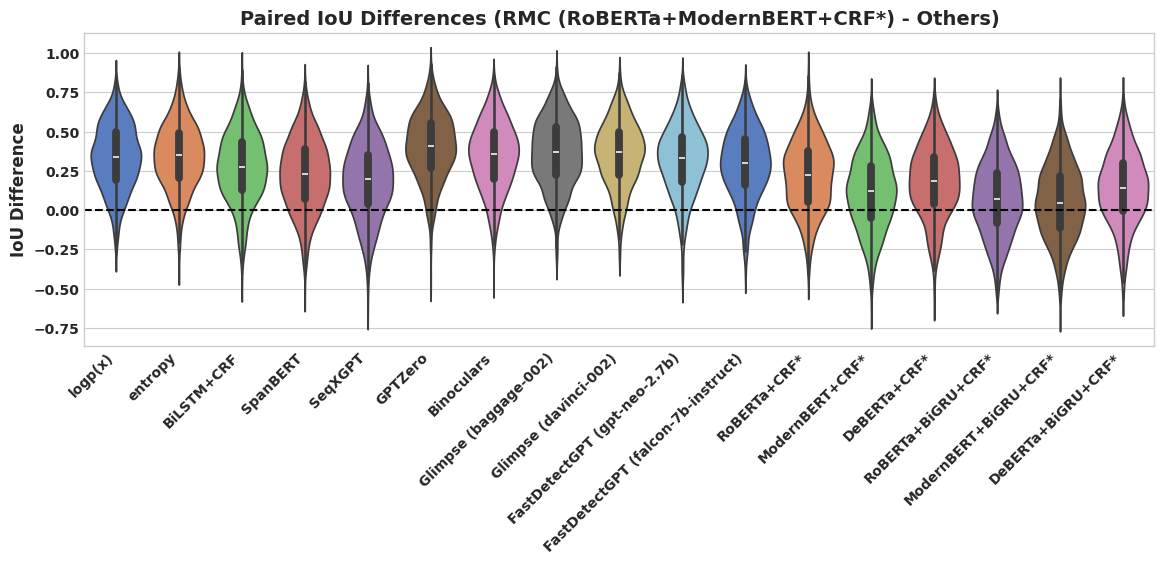}
\caption{Violin plots of paired IoU score differences, showing RMC’s consistent performance superiority over other models.}
\label{fig:iou_diff}
\end{figure}

According to Table~\ref{tab:sbda_sp_results}, We came to a conclusion that the existing zero-shot methods or the statistical based detections are not completely reliable, as upon inclusion of syntactical adversarial attacks the detection of AI-spans was difficult. While our proposed models outperformed all the baselines with a huge difference showcasing significant performance, the highest with 45.75\% SBDA and 41.43\% SP at a threshold of 0.3 respectively by the model \textbf{\textit{RoBERTa + ModernBERT + CRF}}, Figure~\ref{fig:iou_cdf}, \ref{fig:iou_diff}, \ref{fig:combined_iou_distributions}. 
In the Table~\ref{tab:traditional_metrics}, we provided the traditional metrics like Accuracy, Precision, Recall and F1-score which are calculated by the token prediction labels (0-human or 1-AI). The highest values are scored by the same model as above, are 98.28\% across all metrics, accuracy, precision, recall and f1-score. The significance of Info-Mask and the difference between the Traditional Attention and the Info-Mask Mechanism is given below. To assess calibration, we applied temperature scaling on the token-level logits. We observed a marked improvement: \textit{Expected Calibration Error} (\textit{ECE}) dropped from 0.1317 to 0.0036, and the \textit{Brier score} improved from 0.0242 to 0.0029. These results confirm that the model’s confidence estimates become highly reliable after calibration, Figure~\ref{fig:rel_before}, \ref{fig:rel_after}.\\

\begin{table}[h!]
    \centering
    \resizebox{\linewidth}{!}{%
    \begin{tabular}{lcc}
    \toprule
    \textbf{Feature} & \textbf{Traditional Attention} & \textbf{Info-Mask} \\
    \midrule
    Attention basis  & Semantic similarity              & Stylometric divergence \\
    Learned weights  & Via dot-product attention        & Via MHA over style features \\
    Output form      & Context vectors                  & Scalar mask over tokens \\
    Interpretability & Limited (opaque heads)           & Explicit saliency map (per token) \\
    Robustness goal  & Model coherence                  & \textit{Robust attribution} under attack \\
    Role             & Focus across context             & Gate encoder by style attribution \\
    \bottomrule
    \end{tabular}
    }
    \caption{Differences between Traditional Attention and Info-Mask Mechanism}
    \label{tab:diff_info_traditional}
\end{table}

\noindent \textbf{Significance of Info-Mask:} Info-Mask, is significantly different from the Traditional Attention mechanism\cite{vaswani2017attention} and provides explicit interpretability with the help of stylometric features of the text under adversarial attacks, the basic differences are given in Table~\ref{tab:diff_info_traditional}. Based on an example from Figure~\ref{fig:combined-span-analysis}, \ref{fig:infomask_strength}, \ref{fig:attxinfomask}, and \ref{fig:styloheatmap} represent: Token-wise Signal Strength from word to word, where the density or strength is higher for the AI spans and lower for the Human spans; Fused Influence from the Attention and Info-Mask for each word in the sentence, caused by both mechanisms; and a Stylometric Features Heatmap across the words that are used to build the Info-Mask, demonstrating its significance at the word or token level. These collectively enable human-interpretable attribution.

\section{Ablation Study}
\label{ablation}
We performed ablation studies with the best-performing model, \textbf{RMC (RoBERTa + ModernBERT + CRF)}\textsuperscript{*}, covering the impact of optimizations, individual feature inclusion, unseen attacks, unseen generator, and individual attack types.

\begin{enumerate}
    \item \textbf{Ablation with Info-mask and Optimizations techniques:} 1) RMC + No optimizations, 2) RMC + All optimizations, 3) RMC\textsuperscript{*} + No optimizations, and 4) RMC\textsuperscript{*} + All optimizations.
   \item \textbf{Ablation on model performance with Individual Feature:} RMC\textsuperscript{*}-p: Perplexity, RMC\textsuperscript{*}-q: POS Tags, RMC\textsuperscript{*}-r: Punctuation Density, RMC\textsuperscript{*}-s: Lexical Diversity and RMC\textsuperscript{*}-t: Readability.
   \item \textbf{Ablation on model performance for Unseen Attacks:} 1) RMC\textsuperscript{*}-1 + Train with data including only `All Mixed' attack + Test on data with remaining all individual attacks, and 2) RMC\textsuperscript{*}-2 + Train with data excluding only `All Mixed' attack + Test on data with only `All Mixed' attacks.
   \item \textbf{Ablation on model performance for individual Attacks:} RMC\textsuperscript{*} + 1) Mispelling, 2) Character Substitution, 3) Invisible Character, 4) Punctuation Swap, and 5) Upper-Lower Swap
   \item \textbf{Ablation on model performance for unseen-generator:} 1) RMC\textsuperscript{*}-a + Train: (GPT-4o, GPT-4.1), Test: Deepseek, 2) RMC\textsuperscript{*}-b + Train: (Deepseek, GPT-4o), Test: GPT-4.1, 3) RMC\textsuperscript{*}-c + Train: (Deepseek, GPT-4.1), Test: GPT-4o.
   \item \textbf{Ablation with Masking or Gating Variants:} 1) RMC-i: Gradient-Based Attribution masking, 2) RMC-ii: Gating without sylometrics, 3) RMC-iii: Uncertainty-Based Gating, 4) RMC-iv: Purification-Based Masking, and 5) RMC-v: Attention Based Gating.
\end{enumerate}

\begin{table*}[ht]
\centering
\scriptsize
\begin{tabular}{lcccccc}
\toprule
\textbf{\shortstack{Model\\Optimizations}} & \textbf{\shortstack{SBDA\\$\tau@0.3$}} & \textbf{\shortstack{Seg Pre\\$\tau@0.3$}} & \textbf{\shortstack{Train Time\\(sec/epoch)}} & \textbf{\shortstack{Latency\\(ms/1k tok)}} & \textbf{\shortstack{Memory\\(GB)}} & \textbf{\shortstack{Rel.\\Overhead}} \\
\midrule
RMC + No Opt & 0.39 & 0.26 & 14,418 (4hr) & 148 & 12.6 & --- \\
RMC + All Opt & 0.42 & 0.36 & 6,944 (1.9hr) & 115 & 12.1 & $-4\%$ \\
RMC\textsuperscript{*} + No Opt & 0.44 & 0.39 & 18,619 (5.3hr) & 189 & 13.8 & $+10\%$ \\
RMC\textsuperscript{*} + All Opt & \textbf{0.45} & \textbf{0.41} & 8,936 (2.5hr) & 132 & 13.0 & $+3\%$ \\
\bottomrule
\end{tabular}
\caption{Ablation with all optimizations and average training/inference cost for RMC (RoBERTa-base + ModernBERT-base + CRF). Training time is reported per epoch on a single L40S GPU. Inference latency is averaged over 1k tokens (batch=1). Memory is peak GPU usage during inference.}
\label{tab:ablation_opt}
\end{table*}

\noindent We first observe that using all optimization methods provides significant gains in both segmentation and efficiency, decreasing training time and latency with a slight reduction in memory overhead (Table~\ref{tab:ablation_opt}). We note that adding or removing various attacks in training affects generalization, and optimal performance is achieved with all types of attacks included (Table~\ref{tab:unified_ablation}-Top). Second, cross-generator experiments show that RMC\textsuperscript{*} generalizes well to novel text generators and maintains competitive performance even when one source is dropped from training (Table~\ref{tab:unified_ablation}-Upper-Middle). Third, involving individual stylistic features in the Info-Mask demonstrates that lexical variety and POS tags are the most significant contributors, although the collective use of all features provides the highest overall performance (Table~\ref{tab:unified_ablation}-Lower-Middle). Fourth, investigating masking and gating variants identifies an explicit hierarchy, where Info-Mask performs better than all gradient/masking variants, followed by RMC-iv, RMC-iii, RMC-v, RMC-ii, and RMC-i in a declining order of performance (Table~\ref{tab:unified_ablation}-Bottom). Lastly, testing on each type of attack reinforces consistent performance across perturbations, with misspellings being the least damaging and invisible characters being the most difficult (Table~\ref{tab:agreement_attack_single}-Right).

\begin{table*}[ht]
\centering
\scriptsize
\begin{tabular}{llcccccccc}
\toprule
\multirow{2}{*}{\textbf{Ablation}} & \multirow{2}{*}{\textbf{Model}} & \multicolumn{4}{c}{\textbf{SBDA@$\tau$}} & \multicolumn{4}{c}{\textbf{Seg Pre@$\tau$}} \\
\cmidrule(lr){3-6} \cmidrule(lr){7-10}
& & \textbf{0.3} & \textbf{0.5} & \textbf{0.7} & \textbf{0.9} & \textbf{0.3} & \textbf{0.5} & \textbf{0.7} & \textbf{0.9} \\
\midrule
\multicolumn{10}{c}{\textbf{Ablation Study with Inclusion/Exclusion of Attacks}} \\
\midrule
\multirow{2}{*}{\shortstack{Unseen\\Attacks}}
& RMC\textsuperscript{*}-1  & 0.33 & 0.32 & 0.30 & 0.26 & 0.29 & 0.28 & 0.26 & 0.21 \\
& RMC\textsuperscript{*}-2  & 0.42 & 0.41 & 0.39 & 0.34 & 0.38 & 0.37 & 0.35 & 0.30 \\
\midrule
\multicolumn{10}{c}{\textbf{Ablation: RMC\textsuperscript{*} with Held-Out Generators}} \\
\midrule
\multirow{3}{*}{\shortstack{Unseen\\Generator}}
& RMC\textsuperscript{*}-a & 0.40 & 0.37 & 0.35 & 0.30 & 0.36 & 0.34 & 0.31 & 0.27 \\
& RMC\textsuperscript{*}-b & 0.43 & 0.41 & 0.39 & 0.35 & 0.38 & 0.37 & 0.35 & 0.32 \\
& RMC\textsuperscript{*}-c & 0.44 & 0.42 & 0.40 & 0.36 & 0.39 & 0.38 & 0.37 & 0.33 \\
\midrule
\multicolumn{10}{c}{\textbf{Ablation: RMC\textsuperscript{*} with Individual Style Features}} \\
\midrule
\multirow{5}{*}{\shortstack{Individual\\Features}}
& RMC\textsuperscript{*}-p  & 0.37 & 0.36 & 0.34 & 0.30 & 0.33 & 0.32 & 0.30 & 0.25 \\
& RMC\textsuperscript{*}-q  & 0.38 & 0.37 & 0.35 & 0.31 & 0.34 & 0.33 & 0.31 & 0.26 \\
& RMC\textsuperscript{*}-r  & 0.36 & 0.35 & 0.33 & 0.29 & 0.32 & 0.31 & 0.29 & 0.24 \\
& RMC\textsuperscript{*}-s  & 0.39 & 0.38 & 0.36 & 0.32 & 0.35 & 0.34 & 0.32 & 0.27 \\
& RMC\textsuperscript{*}-t  & 0.36 & 0.35 & 0.34 & 0.30 & 0.33 & 0.32 & 0.30 & 0.25 \\
\midrule
\multicolumn{10}{c}{\textbf{Ablation: RMC with Other Masking Variants}} \\
\midrule
\multirow{5}{*}{\shortstack{Masking, \\or Gating\\Variants}}
& RMC-i     & 0.34 & 0.33 & 0.31 & 0.27 & 0.30 & 0.29 & 0.27 & 0.22 \\
& RMC-ii    & 0.36 & 0.35 & 0.33 & 0.29 & 0.32 & 0.31 & 0.29 & 0.24 \\
& RMC-iii   & 0.40 & 0.39 & 0.37 & 0.33 & 0.36 & 0.35 & 0.33 & 0.28 \\
& RMC-iv    & 0.42 & 0.41 & 0.39 & 0.35 & 0.38 & 0.37 & 0.35 & 0.30 \\
& RMC-v     & 0.38 & 0.37 & 0.35 & 0.31 & 0.34 & 0.33 & 0.31 & 0.26 \\
\midrule
& RMC\textsuperscript{*} + All & \textbf{0.45} & \textbf{0.44} & \textbf{0.42} & \textbf{0.37} & \textbf{0.41} & \textbf{0.40} & \textbf{0.38} & \textbf{0.33} \\
\bottomrule
\end{tabular}
\caption{Ablation Studies: (Top) Effect of including/excluding attacks during training, (Top-Middle) Performance with held-out generators, (Bottom-Middle) Effect of individual style features in Info-Mask and (Bottom) Performance with different gradient/masking variants. \textbf{All} represent the model with Info-Mask with all features and trained on complete data with all attacks and generators.}
\label{tab:unified_ablation}
\end{table*}

\begin{table}[ht]
\centering
\scriptsize
\begin{tabular}{c c}
\begin{tabular}{lc}
\toprule
\textbf{Metric} & \textbf{Score} \\
\midrule
ICC & 0.81 \\
Cohen's $\kappa$ & 0.68 \\
Pearson & 0.82 \\
Kendall's $\tau$ & 0.80 \\
Krippendorff's $\alpha$ & 0.79 \\
\bottomrule
\end{tabular}
&
\begin{tabular}{lcc}
\toprule
\textbf{\shortstack{Individual\\Attack}} & \textbf{\shortstack{SBDA\\@0.3}} & \textbf{\shortstack{SP\\@0.3}} \\
\midrule
Misspelling      & 0.36 & 0.28 \\
Character Sub    & 0.33 & 0.24 \\
Invisible Char   & 0.32 & 0.23 \\
Punctuation      & 0.31 & 0.27 \\
Upper-Lower      & 0.33 & 0.29 \\
\bottomrule
\end{tabular}
\end{tabular}
\vspace{-5pt}
\caption{Inter-rater agreement metrics (left) and RMC\textsuperscript{*} performance under attack types (right), also Figure~\ref{fig:heatmap_model_attack}.}
\label{tab:agreement_attack_single}
\end{table}
\vspace{-10pt}

\section{Human Evaluation and Analysis}
\label{appendix:human_eval}
We randomly selected 500 samples and asked 2 individual annotators to evaluate two aspects of the model's output: the quality of its boundary predictions and the usefulness of its \textit{Human-Interpretable Interpretations (HIA)}, both on a 5-point Likert scale (1 to 5). To assess the inter-rater agreement, we computed the following reliability metrics: (1) Intra-class Correlation Coefficient (\textit{ICC}), (2) Cohen’s Kappa, (3) Pearson Correlation, (4) Kendall’s Tau, and (5) Krippendorff’s Alpha as shown in Table~\ref{tab:agreement_attack_single}-Left. We also validated Info-Mask saliencies via a perturbation-based faithfulness check (top-k masking) and found substantial performance degradation ($\Delta$SBDA $-0.12$) when high-mask tokens were removed, but minimal effect ($<$0.01) when low-mask tokens were perturbed. Annotators showed strong agreement ($\kappa$=0.68) and reported reduced decision time when using Info-Mask overlays, suggesting both faithfulness and actionability. The complementary aspects of faithfulness under perturbation and annotator usefulness of Info-Mask can be seen in Table~\ref{tab:faithfulness} and~\ \ref{tab:usefulness}, respectively.

\section{Conclusion and Future Work}
In an effort to avoid detection, some intentionally perturb the generated text using various adversarial strategies. However, the detectability in these cases by the existing detectors struggle to identify such obfuscated AI-generated content with high accuracy as seen in the results. To address this, we introduce a new method that  detects AI-generated spans at a fine-grained level by leveraging a Soft Attribution Masking mechanism, which we term \textit{Info-mask}. This approach not only improves detection robustness in the syntactical adversarial attacks but also provides interpretability through \textit{Human-Interpretable Attribution}, that enable reviewers to understand why specific token sequences are labeled as human- or AI-generated. As a future work, we aim to enhance generalizability of our method to handle complex multimodal content in speech, images etc, and explore multilingual detection capabilities, and integrate real-time feedback systems to support interactive human review.

\section*{Limitations}
While our Info-Mask method enhances fine-grained detection and interpretability, it has a few limitations. It is tailored only for English and may not generalize well to low-resource languages without adaptation. Performance may drop in highly stylized or domain-specific texts where human and AI writing styles overlap. Although interpretability is provided, it still demands reviewer expertise to fully grasp attribution cues. Additionally, the method introduces computational overhead during training and inference, which impact deployment in resource-constrained settings.

\bibliography{custom}

\appendix

\section{More on Data Creation}
\label{appendix:data_creation}
During the construction of the \textit{MAS} dataset, \textit{human-authored texts} were sourced from several well-established benchmark datasets to ensure domain diversity and linguistic quality. Specifically, human Reddit texts were taken from \textsc{\textit{M4}}-\textit{Reddit}~\cite{wang2023m4}, \textsc{\textit{MAGE}}-\textit{YELP}, and \textsc{\textit{MAGE}}-\textit{CMV}~\cite{li-etal-2024-mage}. Human-authored news articles were obtained from the \textsc{\textit{XSUM}} dataset~\cite{narayan-etal-2018-dont}. Wikipedia-based human texts were sourced from \textsc{\textit{M4}}-Wiki and \textsc{\textit{MAGE}}-\textit{SQuAD}, while scientific abstracts from \textsc{\textit{ArXiv}} were collected from \textsc{\textit{MAGE}}-\textit{SciGen}. Finally, question-answering texts were taken from \textsc{\textit{MAGE}}-\textit{ELI5}. \\

\paragraph{Annotation Reliability.} 
The authorship boundaries in the MAS dataset are annotated at the sentence level using explicit span tags: \texttt{<AI\_Start>} and \texttt{</AI\_End>}. These tags denote the beginning and end of AI-generated content within a sentence. Although the annotations are sentence-level, the model performs token-level predictions for finer-grained segmentation. During dataset construction, automatic verification was used to ensure annotation reliability: a sentence was flagged as AI-authored if parsing encountered the \texttt{<AI\_Start>} tag, and remained flagged until the corresponding \texttt{</AI\_End>} tag was reached. This automated span labeling approach guarantees consistent and reproducible annotations throughout the corpus. For the creation of this dataset, we utilized one open-sourced language model, \texttt{DeepSeek-V3-671B}, and three closed-source models: \texttt{GPT-4o}, \texttt{GPT-4.1}, and \texttt{GPT-4.1-mini}.

\paragraph{Split Hygiene.} 
To prevent any potential leakage across train/validation/test sets, we applied strict split hygiene controls: (i) all human-written texts in the training set come from sources disjoint from those in the test set (temporal and article-level separation when the same corpus was used), (ii) we removed overlaps across splits using document-level and $n$-gram similarity filtering (discarding any pair with more than 30\% shared $n$-grams), and (iii) LLM-generated outputs were partitioned such that no generations from the same LLM instance or sampling pool appear in both training and test. These measures ensure that performance cannot be attributed to memorization or leakage but reflects genuine generalization ability.

\section{Justification on Attack Selection:}
\label{sec:adv_attacks}

In this work, we focus primarily on \textbf{syntactical adversarial attacks}, as these occur more naturally in the collaborative text settings of human-AI in the real world. The following six perturbations are considered in our experiments:

\begin{enumerate}
    \item \textbf{Misspelling attack:} This perturbation introduces typographical errors by altering characters in a word, that shows a realistic human-like mistakes. These perturbations can change the way tokenization is performed performed and can affect model predictions without changing semantics.
    \item \textbf{Character Substitution attack:} It replaces characters with visually similar Unicode counterparts. Substitutions like this preserve the appearance but alter the underlying token IDs, and can mislead models.
    \item \textbf{Invisible Character attack:} Zero-width or non-printing Unicode characters are inserted within words, making them visually identical to humans. These insertions disrupt token boundaries and can lead challenges to sub-word tokenizers.
    \item \textbf{Punctuation substitution attack:} This perturbation replaces a standard punctuation with another punctuation. This exploits tokenizer reliance on punctuation as structural cues in text.
    \item \textbf{Upper-Lower Swap attack:} This attack randomly alters the case of characters within a word. While humans easily interpret these case variations, but models often rely on the consistent casing for token semantics.
    \item \textbf{All mixed attack:} A mixture perturbation that combines all the above attacks within the same input. It increases robustness evaluation difficulty by simulating more complex, real-world adversarial scenarios.
\end{enumerate}

To identify the most used attack types by the people, we conducted a user study involving \textbf{213} participants, including undergraduate students familiar with AI-generated content. A total of \textbf{10} syntactical adversarial perturbation types namely \textit{All Mixed, Misspelling, Character Substitution, Invisible Character, Punctuation Substitution, Upper-Lower Swap, Whitespace Jittering, Homoglyph Insertion, Emoji Appending, Leetspeak Transformation (LST)}, were presented in mixed-text settings, and participants were asked to flag attacks (can choose multiple) based on their usage frequency for undetectability. The results of the survey showed that the above six attacks had the highest counts in terms of being chosen as the most difficult to detect. The results of the count chosen by the students in the survey are visualized in Figure \ref{fig:survey_res}.

\begin{figure*}[ht]
    \centering
    \includegraphics[width=0.8\textwidth]{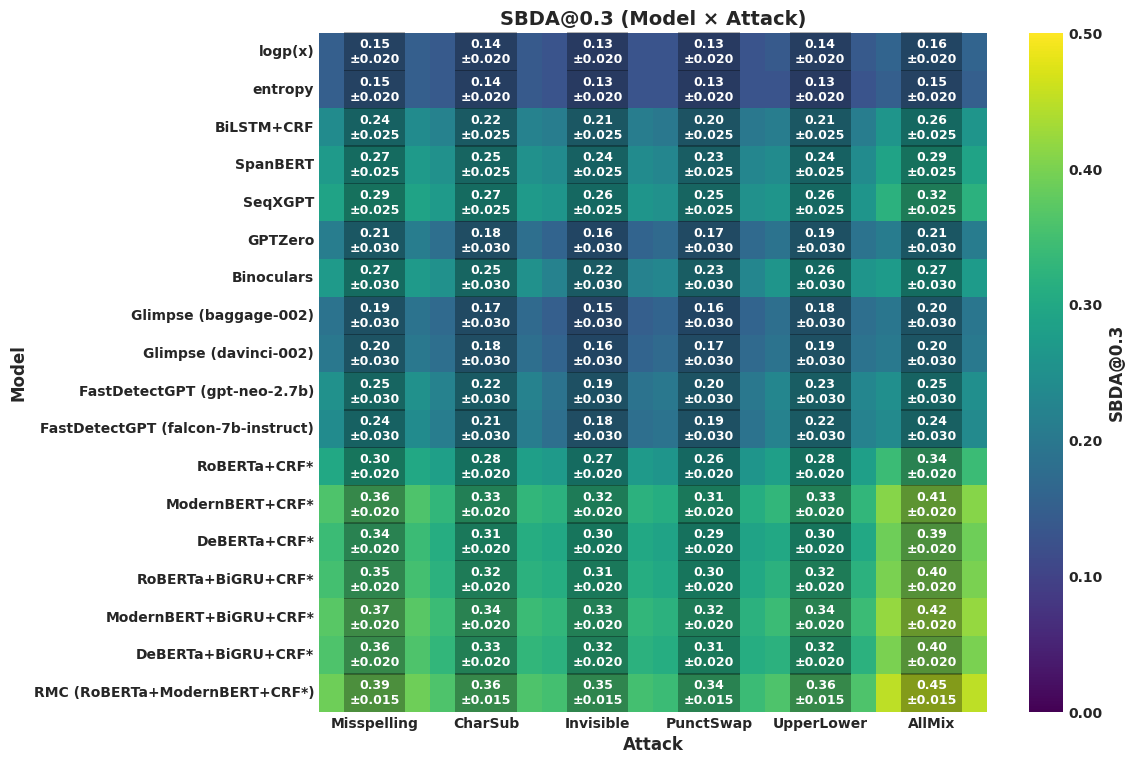}
    \caption{Heatmap with Confidence Interval(CI) of SBDA@0.3 scores across various adversarial attacks. Brighter yellow indicates higher scores, showcasing the superior and consistent robustness of our RMC model.}
    \label{fig:heatmap_model_attack}
\end{figure*}

\begin{figure*}[ht]
\centering
\begin{subfigure}{0.48\textwidth}
    \begin{tikzpicture}
    \scriptsize
    \begin{axis}[
        width=\linewidth,
        ybar,
        bar width=0.5cm,
        enlarge x limits=0.2,
        legend style={at={(0.5,-0.2)}, anchor=north, legend columns=-1},
        ylabel={Time (Hours)},
        xlabel={Model},
        symbolic x coords={RMC + No Opt, RMC + All Opt, RMC* + No Opt, RMC* + All Opt},
        xtick=data,
        x tick label style={rotate=30, anchor=east},
        ymin=0,
        ymax=6,
        ymajorgrids=true,
        yticklabel={\pgfmathparse{\tick}\pgfmathprintnumber{\pgfmathresult}},
        nodes near coords,
        nodes near coords align={vertical},
        every node near coord/.append style={font=\tiny, yshift=0.0ex},
        point meta=explicit,
        title={Avg. Training Time for each Epoch},
    ]
    \addplot coordinates {
        (RMC + No Opt, 4) [14418]
        (RMC + All Opt, 1.9) [6944]
        (RMC* + No Opt, 5.3) [18619]
        (RMC* + All Opt, 2.5) [8936]
    };
    \end{axis}
    \end{tikzpicture}
    \caption{Avg. Training time for each Epoch in seconds (y-axis labeled in hours) for the models with and without the inclusion of Optimization Techniques.}
    \label{fig:train_epoch_time}
\end{subfigure}\hfill
\begin{subfigure}{0.48\textwidth}
    \begin{tikzpicture}
    \begin{axis}[
        width=\linewidth,
        ybar,
        bar width=10pt,
        ylabel={Number of Votes},
        symbolic x coords={
            All Mixed, Misspelling, Char Subs,
            Invisible, Punctuation,
            Upper-Lower, Whitespace,
            Homoglyph, Emoji, LST
        },
        xtick=data,
        xticklabel style={rotate=45, anchor=east},
        ymin=0,
        ymax=120,
        nodes near coords,
        enlarge x limits=0.08,
        every node near coord/.append style={font=\small},
        title={Votes of participants on each Attack.},
    ]
    \addplot coordinates {
        (All Mixed,108)
        (Misspelling,92)
        (Char Subs,86)
        (Invisible,80)
        (Punctuation,77)
        (Upper-Lower,69)
        (Whitespace,34)
        (Homoglyph,29)
        (Emoji,24)
        (LST,19)
    };
    \end{axis}
    \end{tikzpicture}
    \caption{Survey Results: Counts of Syntactic Adversarial Attacks chosen by the students in the survey.}
    \label{fig:survey_res}
\end{subfigure}

\end{figure*}

\section{Features for Info-Mask}
\label{sec:info_mask_features}
We selected the following stylistic features: 1) Perplexity: the unpredictability of a token under a language model; 2) POS tag density: frequency of each computed using all POS categories to capture syntactic composition; 3) Punctuation Density: proportion of punctuation marks relative to total tokens; 4) Lexical Diversity:  the richness of vocabulary through the ratio of unique words to total words; and 5) Readability: quantifying textual clarity based on sentence length and word complexity. They all combinedly capture both surface-level and writing characteristics. These features are lightweight, interpretable, and well-established in stylometry and authorship attribution that help distinguish between human and LLM-generated styles. Their complementary nature supports Info-Mask’s token-level features in guiding authorship-consistent segmentation.

\section{Derivation of Info-Mask}
\label{appendix:info_mask_derivation}
To incorporate style-sensitive control into token representations, each token’s hand-crafted style feature vector $\mathbf{s}_i \in \mathbb{R}^f$ is first projected into a latent representation space using a linear transformation followed by a non-linear activation: $\mathbf{v}_i = \text{ReLU}(\mathbf{W}_s \cdot \mathbf{s}_i + \mathbf{b}_s)$. Here, $\mathbf{W}_s \in \mathbb{R}^{d_s \times f}$ is a learnable weight matrix that transforms the $f$-dimensional style feature vector into a $d_s$-dimensional latent space, and $\mathbf{b}_s \in \mathbb{R}^{d_s}$ is the bias term. The output $\mathbf{v}_i$ is the style-projected vector corresponding to the $i$-th token. Taken together, the representations for all the tokens form a matrix $\mathbf{V} = [\mathbf{v}_1, \mathbf{v}_2, ..., \mathbf{v}_T] \in \mathbb{R}^{T \times d_s}$, where $T$ is the sequence length. Next, a self-attention mechanism is applied to the style-projected matrix $\mathbf{V}$ using multi-head attention to capture contextual dependencies among style features: $\mathbf{A} = \text{MultiHead}(\mathbf{V})$. The output $\mathbf{A} \in \mathbb{R}^{T \times d_s}$ contains attention-enhanced representations of the style signals. From this, a scalar mask value $m_i \in [0,1]$ is derived for each token $i$ by summing the attention scores across the feature dimension and passing the result through a sigmoid function: $m_i = \sigma\left( \sum_{j=1}^{d_s} A_{i,j} \right)$. This scalar $m_i$ acts as a soft gating mask to determine how much of the original token representation $\mathbf{z}_i$ should be remain same or retained. Finally, the Info-Mask is applied by element-wise scaling of the original encoder output $\mathbf{z}_i$ with the mask $m_i$: $\tilde{\mathbf{z}}_i = m_i \cdot \mathbf{z}_i$. This derives a style-aware representation $\tilde{\mathbf{z}}_i$ where stylistically irrelevant or uninformative tokens are down-weighted, and style-relevant tokens are preserved or emphasized, and guides downstream model with interpretable, feature-informed attention.

\section{CRF Loss}
\label{appendix:crf_loss}
During the training, the loss function is the CRF loss, which is a negative log-likelihood and computed in the following way: Let the model output at each time step be a vector of size L (the number of possible labels). Collectively, the emissions across the entire sequence of length T are represented as: $\mathbb{O} = [o_1, o_2, .., o_T] \in \mathbb{R}^{T \times L}$, where, each $o_i$ is calculated as $\mathbf{o}_i = \mathbf{W}_c \cdot \mathbf{u}_i + \mathbf{b}_c$ and the trainable transition matrix $\mathbb{T} \in \mathbb{R}^{L \times L}$, where $T_{i,j}$ represents the transition score from label $i$ to label $j$ and. This matrix captures the likelihood of label sequences. Finally the true label sequence is denoted as ${y} = [{y1},..,{y_T}]$ where $y_i \in {0, 1}$.

The score of a label sequence $y$ given the emissions $O$ is the sum of transition scores between consecutive labels and emission scores for the predicted label at each position is given by Equation~\ref{eq:scorer}, where, the first term aggregates transition scores for label transitions from $y_0$ (start) to $y_{T+1}$ (end) and the The second term aggregates the emission scores for the correct label at each token position. Such that, the CRF loss which is the negative loss likelihood will be calculated as the Equation~\ref{eq:loss}, where, $s(O, y)$ is the CRF score  of the correct label sequence which can be calculated using \ref{eq:scorer}, The denominator sums over all possible label sequences $y^`$, making this a global normalization.

\begin{subequations}
\label{eq:crf_eqs}
\begin{minipage}{0.48\linewidth}
    \begin{equation}
    \label{eq:scorer}
    s(\mathbf{O}, \mathbf{y}) = \sum_{t=0}^{T} \mathbf{T}_{y_t, y_{t+1}} + \sum_{t=1}^{T} \mathbf{o}_{t, y_t}
    \end{equation}
\end{minipage}\hfill
\begin{minipage}{0.48\linewidth}
    \begin{equation}
    \label{eq:loss}
    \mathcal{L} = -\left( s(\mathbf{O}, \mathbf{y}) - \log \sum_{\mathbf{y}'} \exp(s(\mathbf{O}, \mathbf{y}')) \right)
    \end{equation}
\end{minipage}
\end{subequations}

\section{More on Comparisons}
\subsection{Experimental Setup}
All experiments were carried out through Amazon Web Services (AWS) EC2 instance optimized for accelerated computing. The instance type is \textbf{g6e.xlarge} that is equipped with 3rd generation AMD EPYC 7R13 processors, 4 virtual CPUs, and 150 GiB of RAM. For GPU acceleration, the setup uses \textbf{NVIDIA L40S} Tensor Core GPU with 48 GB of dedicated memory, enabling efficient handling of large-scale model training and inference tasks. The operating system environment was based on Ubuntu Server 24.04 LTS. We spent $\approx$ USD 720 on AWS EC2 for $\approx$400 hours of GPU utilization.

\subsection{Statistical Based-detectors}
\label{appendix:stat_detect}
\begin{enumerate}
    \item \textbf{\textit{Log p(x)}} serve the \textit{unexpectedness} computes the negative log-likelihood $\log p(x_i|x_{<i})$ for each of the token $x$ that is obtained from the encoder transformer model and $x_i$ is the token position at $i$ in the sequence. Here based on the values of the $log p(x) = \sum_{i=1}^{n}logp(x_i|x_{<i})$ that were computed, we differentiate whether it is in a human part or the AI part. As the log p(x) trends for more AI-like texts have lower values and human-like texts have higher values respectively. Thus, tokens with lower values of $\log p(x_i)$ are flagged as more likely to be AI generated.
    \item \textit{\textbf{Entropy-based}} detection measures the \textit{uncertainty} of the model's next token-prediction. The entropy that is given by $S(p_i) = -\sum_j p_{ij} \log p_{ij}$ will be computed based on the softmax probability distribution, where $p_i$ and $p_{ij}$ are the predicted probability distribution for the next token at position $i$ in the vocabulary and the probability assigned to vocabulary token j at position i respectively. High entropy resembles the presence of AI-part, and we made the calculation as tokens with entropy value having more than the 75th percentile are labeled as likely AI-generated. These statistical signals require no additional training and are entirely model-internal, making them attractive for fast, lightweight zero-shot AI text detection.
\end{enumerate}

\subsection{Zero-shot Detectors}
\label{appendix:zeroshot_detect}
\begin{enumerate}
    \item \textbf{\textit{Fast-DetectGPT}} which detects AI text by the perturbed samples and evaluated texts likelihood under different model prompts ans this has two variations: a) \textit{falcon-7b/falcon-7b-instruct} where falcon-7b is used as the sampling model and falcon-7b-instruct as the scoring model, and b) \textit{gpt-neo-2.7b} where this model itself is used as both sampling and scoring models,

    \item \textbf{\textit{Glimpse}} detects AI text with the help of principle of uncertainty-guided token perturbation, where this have two variations: a) \textit{davinci-002} and b) \textit{baggage-002}, where these models itself are used as a scoring model,

    \item \textbf{\textit{Binoculars}}, that performs fine-grained AI text detection by comparing representations between base- and instruction-tuned language models.

    \item \textbf{\textit{GPTZero}} detector uses simple statistical signals such as perplexity and burstiness which shows the best variation in sentence lengths and structures for distinguishing human-written and AI-generated text parts. It is designed for fast and interpretable predictions, and the most useful in educational and content moderation settings.
\end{enumerate}

\subsection{Other Gradient and Masking Variants}

\begin{enumerate}
    \item \textbf{RMC-i: Gradient-Based Attribution Masking} is created through gradients' feature important scores which are \textit{Integrated Gradients}, which are used as soft weights over hidden states. This provides robust interpretability in the form of saliency maps but is prone to gradient noise,  and has expensive computation overhead, and susceptibly to specially crafted gradient-based attacks.
    \item \textbf{RMC-ii: Gating without Stylometrics} It uses MLP over the hidden states to learn soft gates directly from context patterns, trained end-to-end. This is inefficient and architecture-multiplicative but clear, less explainable, and vulnerable to syntactic attacks because it's excessive reliance on semantics.
    \item \textbf{RMC-iii: Uncertainty-Based Gating} uses uncertainty estimates (Monte Carlo dropout entropy) to downweight uncertain tokens. Enhances calibration and stability to distributional changes at the cost of modest overhead, but provides limited interpretability and lesser resistance to stylometric attacks.
    \item \textbf{RMC-iv: Purification-Based Masking} It preprocesses inputs with denoising (autoencoders) and computes masks from differences between original and cleansed text. Resilient to adversarial noise but may lose subtle stylistic signals, introduces preprocessing overhead, and degrades precision in boundary detection.
    \item \textbf{RMC-v: Attention-Based Gating} It utilizes transformer self-attention weights as contextual masks, embedded natively with low overhead. Ineffective for capturing semantic and syntactic relationships but still susceptible to style-based perturbations and only offers low-level interpretability.
\end{enumerate}

\subsection{Statistical Significance of Reported Improvements}
To ensure the robustness of our results, we performed a paired t-test on five independent runs with different random seeds. The observed improvement in segment precision at the threshold $\tau$ = 0.3 from 0.26 (RMC + no option) to 0.41 (RMC\textsuperscript{*} + all option) produced a p-value of 0.0082, indicating statistical significance at the confidence level 99\%. Similarly, the improvement in SBDA from 0.39 to 0.45 was also statistically significant, with p $\le$ 0.01. These results support the conclusion that the performance gains introduced by the Info-Mask and optimization techniques are consistent and unlikely to be due to random variation. While calibration, the fitted temperature parameter was 0.1664, selected via validation. This value was used consistently across experiments when reporting post-calibration metrics.

\begin{figure*}[ht]
\centering
\begin{subfigure}{0.48\textwidth}
    \centering
    \includegraphics[width=\linewidth]{Images/reliability_before.png}
    \caption{Reliability diagram of the baseline model before calibration, showing poor calibration with significant deviation from the ideal diagonal.}
    \label{fig:rel_before}
\end{subfigure}\hfill
\begin{subfigure}{0.48\textwidth}
    \centering
    \includegraphics[width=\linewidth]{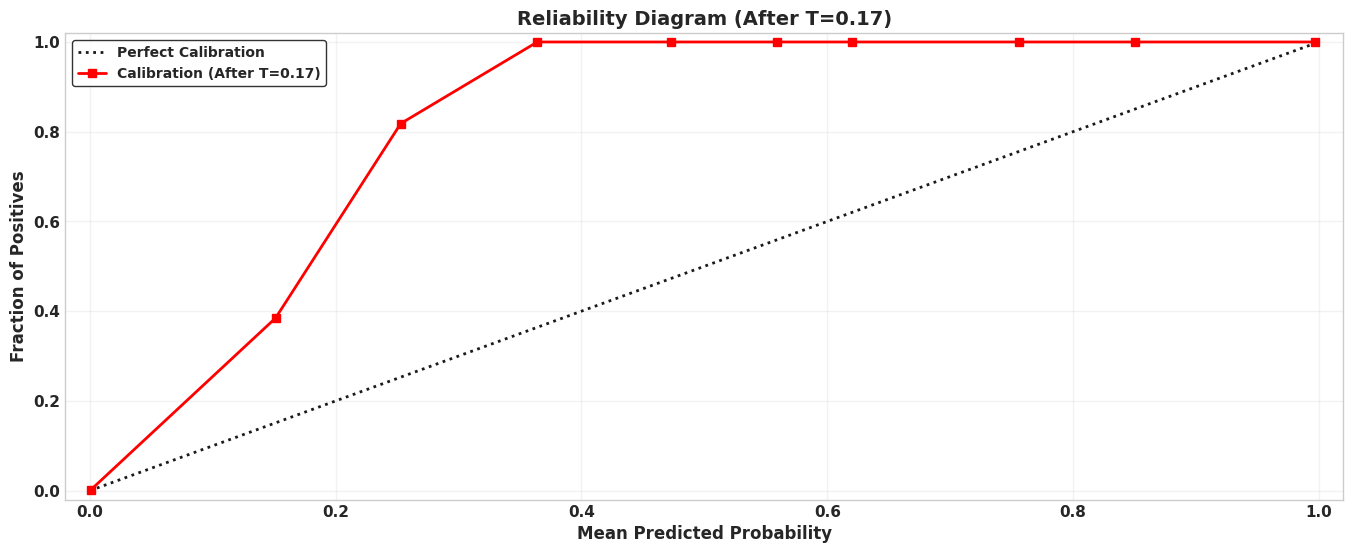}
    \caption{Reliability diagram after temperature scaling (T=0.17), demonstrating improved calibration with the curve closely following the ideal diagonal.}
    \label{fig:rel_after}
\end{subfigure}
\end{figure*}

\begin{table}[ht]
    \centering
    \resizebox{\linewidth}{!}{
    \begin{tabular}{lcccc}
        \toprule
        \textbf{Proposed Approach Model} & \textbf{Acc} & \textbf{Prec} & \textbf{Rec} & \textbf{F1} \\
        \midrule
        RoBERTa + CRF\textsuperscript{*}          & 0.82 & 0.82 & 0.82 & 0.82 \\
        DeBERTa + CRF\textsuperscript{*}          & 0.90 & 0.91 & 0.90 & 0.90 \\
        ModernBERT + CRF\textsuperscript{*}       & 0.92 & 0.93 & 0.92 & 0.93 \\
        RoBERTa + BiGRU + CRF\textsuperscript{*}    & 0.96 & 0.96 & 0.96 & 0.96 \\
        ModernBERT + BiGRU + CRF\textsuperscript{*}   & 0.97 & 0.97 & 0.97 & 0.97 \\
        DeBERTa + BiGRU + CRF\textsuperscript{*}    & 0.97 & 0.97 & 0.97 & 0.97 \\
        RoBERTa + ModernBERT + CRF\textsuperscript{*} & 0.98 & 0.98 & 0.98 & 0.98 \\
        \bottomrule
    \end{tabular}
    }
    \caption{Traditional Evaluation metrics Accuracy, Precision, Recall and F1-Score for the models on proposed approach (with info-mask), where \textsuperscript{*} indicates Info-Mask is included.}
    \label{tab:traditional_metrics}
\end{table}

\subsection{Token vs. Span Discrepancy}  
We note that token-level metrics ($\approx$ 0.98) are substantially higher than span-level scores ($\approx$ 0.45 SBDA, 0.41 SegPre). This discrepancy is expected in span detection tasks: even a small boundary shift of 1–2 tokens causes a span to be counted as incorrect, despite nearly perfect token labeling. To better understand this, we analyzed boundary errors for our best model, \textbf{RMC\textsuperscript{*}}, and report the distribution of errors alongside a boundary-tolerant metric in Tables~\ref{tab:boundary_errors} and \ref{tab:relaxed_metric}. Results confirm that most errors are minor (off by only a few tokens), and that under IoU-based relaxed matching, performance aligns with the strong token-level scores. This demonstrates the practical robustness of our approach despite modest raw span-level numbers.   

\begin{table}[ht]
\centering
\begin{tabular}{lc}
\toprule
\textbf{Error Type} & \textbf{Percentage} \\
\midrule
Off-by-1 token & 46\% \\
Off-by-3 tokens & 25\% \\
Off-by-5 tokens & 15\% \\
Off-by-10 tokens & 8\% \\
Larger ($\ge$10) & 6\% \\
\bottomrule
\end{tabular}
\caption{Boundary error distribution for RMC\textsuperscript{*}.}
\label{tab:boundary_errors}
\end{table}

\begin{table}[ht]
\centering
\begin{tabular}{lc}
\toprule
\textbf{Metric} & \textbf{Score} \\
\midrule
SBDA (strict) & 0.45 \\
SegPre (strict) & 0.41 \\
Relaxed Span Acc. (IoU $\geq$ 0.5) & 0.82 \\
\bottomrule
\end{tabular}
\caption{Relaxed span accuracy (IoU $\geq$ 0.5).}
\label{tab:relaxed_metric}
\end{table}


\begin{table}[ht]
\centering
\scriptsize
\begin{tabular}{lcc}
\toprule
\textbf{Perturbation} & \textbf{SBDA@0.3} & \textbf{SegPre@0.3} \\
\midrule
No perturbation (clean test set)     & 0.47 & 0.43 \\
Mask top-10\% tokens                  & 0.33 & 0.29 \\
Mask bottom-10\% tokens               & 0.46 & 0.42 \\
Shuffle top-10\% tokens               & 0.34 & 0.30 \\
Shuffle bottom-10\% tokens            & 0.46 & 0.41 \\
\bottomrule
\end{tabular}
\caption{Faithfulness via perturbation of Info-Mask tokens (RMC\textsuperscript{*}).  ``No perturbation'' refers to evaluation on the clean test set; perturbation rows show performance after applying the stated modification to the same clean set.}
\label{tab:faithfulness}
\end{table}

\begin{table}[ht]
\centering
\begin{tabular}{lcc}
\toprule
\textbf{Metric} & \textbf{Score} \\
\midrule
Inter-annotator $\kappa$ & 0.68 \\
Decision time (s) w/o mask & 21.4 \\
Decision time (s) w/ mask & 8.7 \\
Usefulness rating (1–5) & 4.3 \\
\bottomrule
\end{tabular}
\caption{Annotation usefulness with Info-Mask overlays.}
\label{tab:usefulness}
\end{table}

\section{Optimization Techniques}
\label{sec:opt_techniques}
As discussed above, to maintain the training stability and reduce effective training period, a few optimizations were utilized. Inclusion of these techniques reflect significantly in the model training time along with the performance as shown in Table \ref{tab:ablation_opt}, while upon including them, the training time for each epoch is reduced with a large difference either with or without the Info-mask, Figure~\ref{fig:train_epoch_time}. 
The importance of each optimization is given below: 
\begin{enumerate}
    \item Layer-wise Learning Rate Decay (LLRD), which progressively reduces learning rates across lower layers to preserve pre-trained representations and in parallel, it allows deeper layers to adapt to downstream tasks;
    \item Dynamic Dropout varies the dropout rate during training period to enhance regularization adaptively based on learning dynamics and this avoids overfitting;
    \item Gradient Clipping is employed to prevent gradient explosion during backpropagation in the pass and stabilize training under adversarial noise; and
    \item Xavier Initialization is utilized to maintain a balanced variance in weights across layers, and such that improving convergence rates in deep transformer architectures.
\end{enumerate}

\section{Data Generation}
\label{appendix:data_gen}
\noindent \textbf{H$\rightarrow$M text Generation:} 
To generate H $\rightarrow$ M transitions, we begin with a fully human-written text and truncate the end segment. The truncated text is then fed to AI model, and asked to complete it in a coherent and contextually relevant in the same tone. This results in a hybrid sample where the beginning is authored by a human and the continuation is generated by a machine.

\noindent \textbf{M$\rightarrow$H text Generation:} 
For M$\rightarrow$H transitions, we follow the inverse process. We start with the same human-written corpus but truncated the first part. The remaining text is used as a target for the model, which is prompted to generate a coherent beginning. This yields a sample where the initial part is machine-generated and the latter part is human-authored.

\noindent \textbf{Completely Mixed-Text Generation:} 
The following prompt is used for the generation of deeply-mixed texts. While, the above setting `H$\rightarrow$M' and `M$\rightarrow$H' are comparatively easy to generate than the generation of Deeply-mixed text generation, where, this setting require much more care is needed in order to maintain the coherence. 

\begin{tcolorbox}[colback=gray!10, colframe=gray, title=Prompt for Deeply Mixed texts Data      Generation, fonttitle=\bfseries]
\scriptsize
You are tasked with creating content for an AI-human collaborative document. The document has missing parts marked by \texttt{<AI\_Start></AI\_End>} tags. Your job is to generate a single novel sentence to fill the gap between \texttt{<AI\_Start>} and \texttt{</AI\_End>}. The sentence should: 
\begin{enumerate}
    \item Be accurate and relevant to the topic implied by the surrounding context, regardless of the domain,
    \item Fit seamlessly with the surrounding text, maintaining the document's flow and style,
    \item Be distinct from any original content, offering a fresh perspective or detail, and 4) Be concise and suitable for sentence segmentation studies.
\end{enumerate}

\textbf{Context before the missing part:} \texttt{"\{left\_context\}"}, 

\textbf{Context after the missing part:} \texttt{"\{right\_context\}"}

Reply with \textbf{ONLY} the sentence to be placed between \texttt{<AI\_Start>} and \texttt{</AI\_End>}, without including the tags themselves.
\end{tcolorbox}



\begin{table}[ht]
\centering
\scriptsize
\resizebox{\linewidth}{!}{
\begin{tabular}{lc}
\toprule
\textbf{Hyperparameter} & \textbf{Value} \\
\midrule
Batch size & 64 \\
Epochs & 5 \\
Weight decay & 0.01 \\
Gradient clipping & 1.0 \\
Initial dropout & 0.1 \\
Warm-up percentage & 0.1 \\
Early stopping patience & 2 \\
Number of labels & 2 \\
Lexical window size & 5 \\
Style feature dimension & 5 \\
Number of attention heads & 5 \\
Style hidden size & 64 \\
Max sequence length & 512 \\
Dynamic dropout range & 0.1 -- 0.3 \\
Optimizer & AdamW \\
Learning rate scheduler & Cosine annealing \\
Learning rate & $1e^{-6} \rightarrow 5e^{-6} \rightarrow 1e^{-5} \rightarrow 1e^{-4}$ \\
\bottomrule
\end{tabular}
}
\caption{Model Hyperparameters}
\label{tab:hyperparameters}
\end{table}

\begin{figure*}[ht]
\centering
\begin{subfigure}{0.32\textwidth}
    \includegraphics[width=\linewidth]{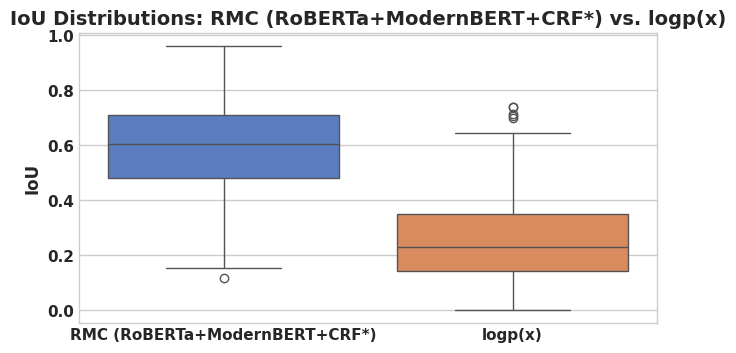}
\end{subfigure}\hfill
\begin{subfigure}{0.32\textwidth}
    \includegraphics[width=\linewidth]{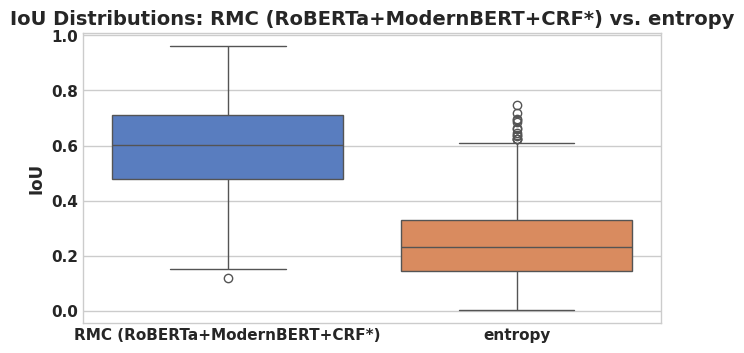}
\end{subfigure}\hfill
\begin{subfigure}{0.32\textwidth}
    \includegraphics[width=\linewidth]{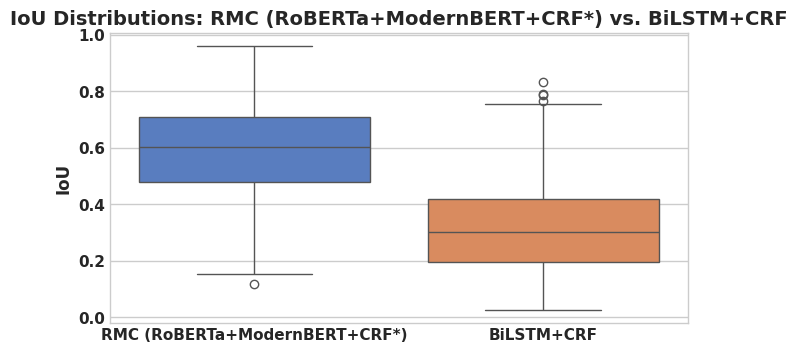}
\end{subfigure}
\vspace{1ex} 
\begin{subfigure}{0.32\textwidth}
    \includegraphics[width=\linewidth]{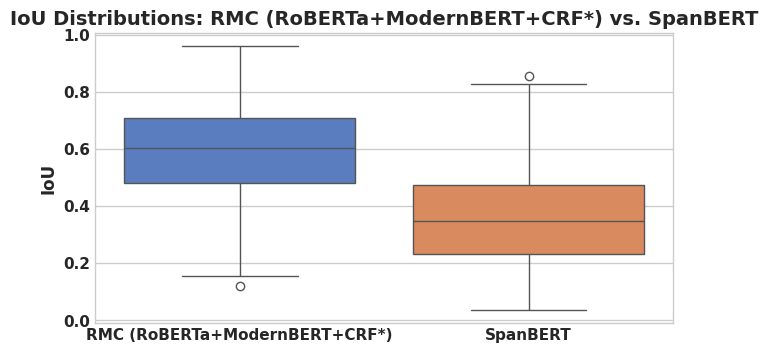}
\end{subfigure}\hfill
\begin{subfigure}{0.32\textwidth}
    \includegraphics[width=\linewidth]{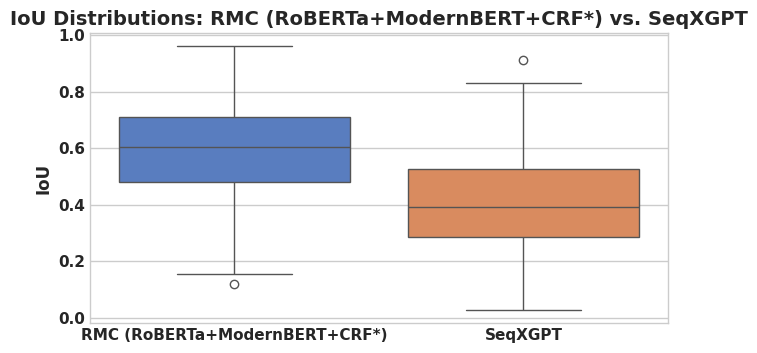}
\end{subfigure}\hfill
\begin{subfigure}{0.32\textwidth}
    \includegraphics[width=\linewidth]{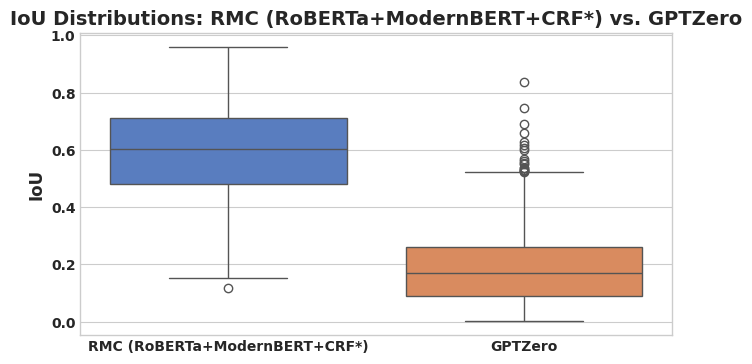}
\end{subfigure}
\vspace{1ex} 
\begin{subfigure}{0.32\textwidth}
    \includegraphics[width=\linewidth]{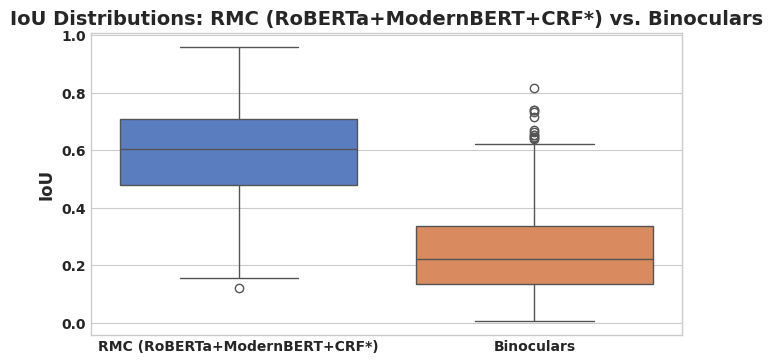}
\end{subfigure}\hfill
\begin{subfigure}{0.32\textwidth}
    \includegraphics[width=\linewidth]{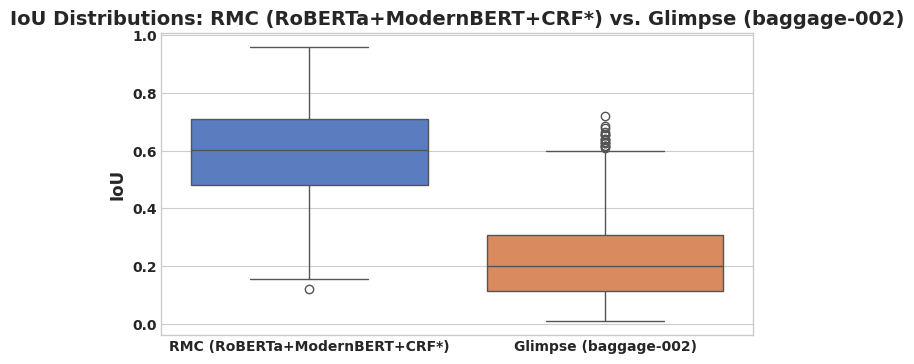}
\end{subfigure}\hfill
\begin{subfigure}{0.32\textwidth}
    \includegraphics[width=\linewidth]{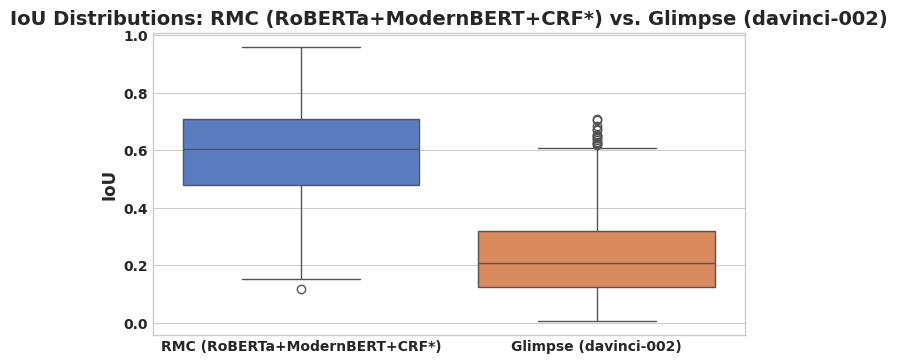}
\end{subfigure}
\vspace{1ex}
\begin{subfigure}{0.32\textwidth}
    \includegraphics[width=\linewidth]{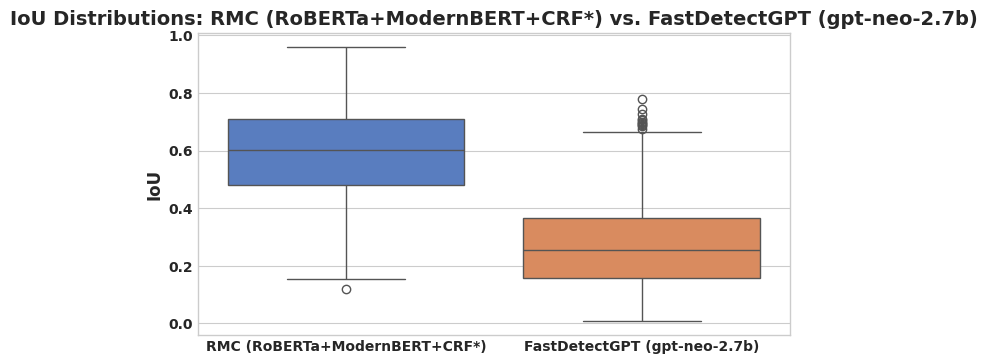}
\end{subfigure}\hfill
\begin{subfigure}{0.32\textwidth}
    \includegraphics[width=\linewidth]{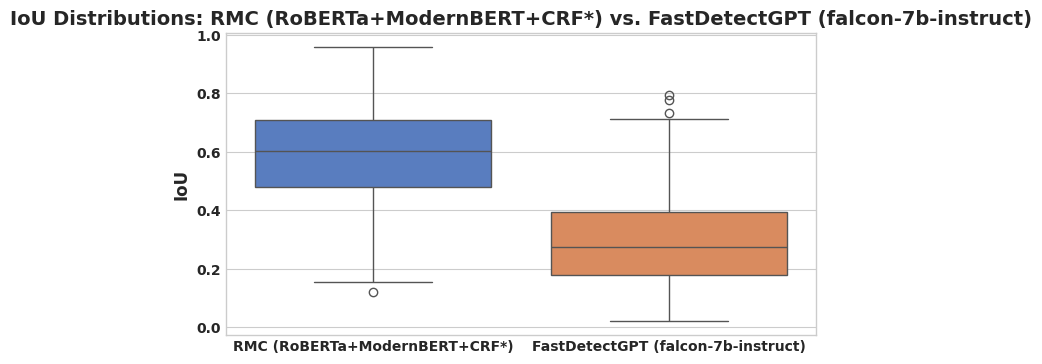}
\end{subfigure}\hfill
\begin{subfigure}{0.32\textwidth}
    \includegraphics[width=\linewidth]{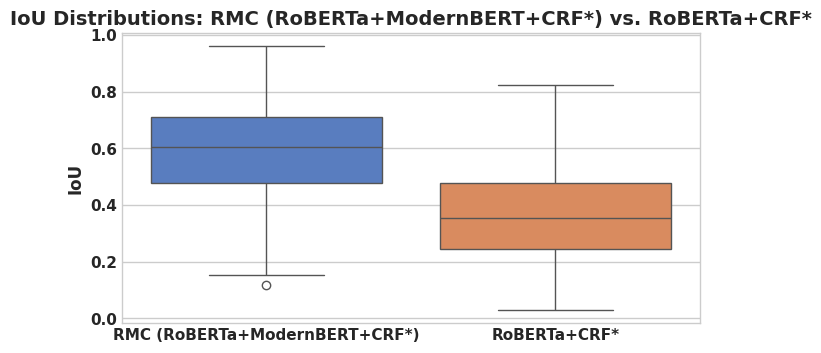}
\end{subfigure}
\vspace{1ex}
\begin{subfigure}{0.32\textwidth}
    \includegraphics[width=\linewidth]{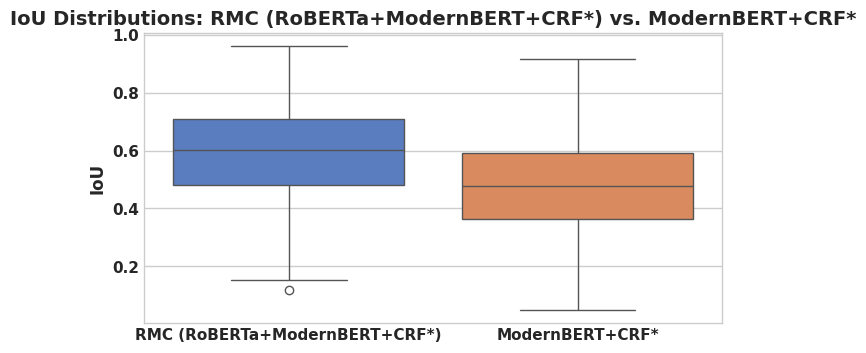}
\end{subfigure}\hfill
\begin{subfigure}{0.32\textwidth}
    \includegraphics[width=\linewidth]{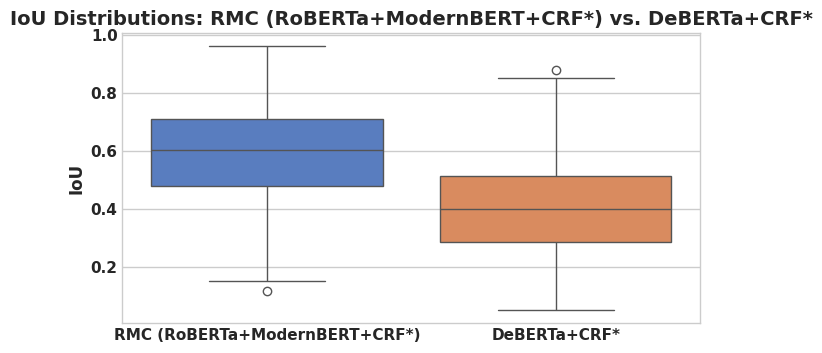}
\end{subfigure}\hfill
\begin{subfigure}{0.32\textwidth}
    \includegraphics[width=\linewidth]{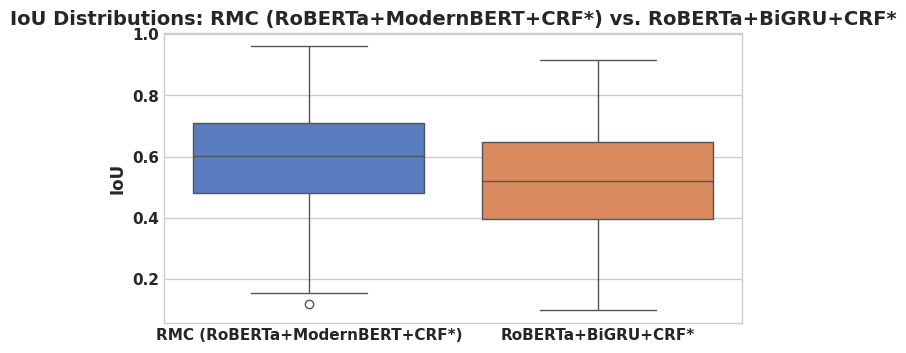}
\end{subfigure}
\vspace{1ex}
\begin{subfigure}{0.32\textwidth}
    \includegraphics[width=\linewidth]{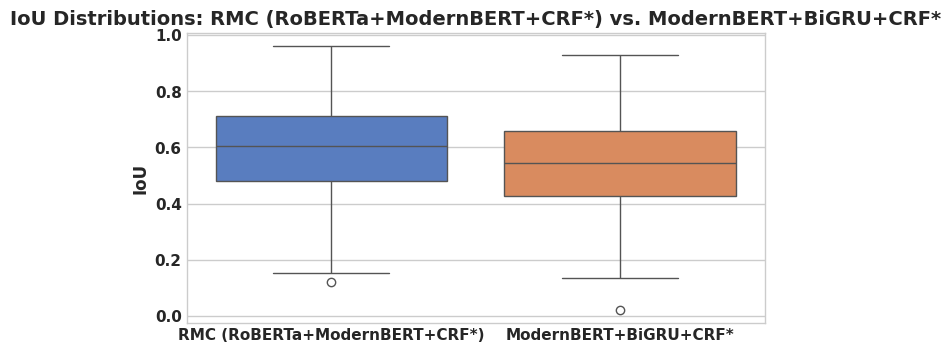}
\end{subfigure}\hfill
\begin{subfigure}{0.32\textwidth}
    \includegraphics[width=\linewidth]{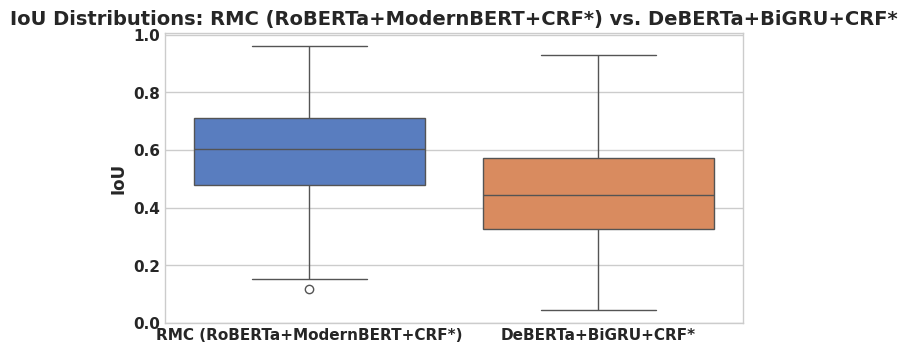}
\end{subfigure}
\caption{IoU Distribution Comparisons with RMC\textsuperscript{*} with all other models.}
\label{fig:combined_iou_distributions}
\end{figure*}

\begin{figure*}[ht]
\centering
\begin{tcolorbox}[colback=white, colframe=black, title=Original vs Predicted Spans, width=\textwidth, boxrule=0.5pt, arc=4pt, auto outer arc]
\textbf{Original Text:}

I don't that the libraries should take awy the books, movies, or any other materials. What if their little kids want to watch a movie or something and they can't because all of the materials are gone. I think that the librarians should keep all the materials. <AI\_Start>If they take away one book, then they have to take away all the other books that someone else might find offensive. This would leave nothing for anyone to read. If someone finds a book, movie, or any other material offensive, they should not take it out. They should not read it or watch it. It's not fair to take away someone else's right to read or watch what they want. Everyone has different opinions and beliefs, and they should be able to express them through their choice of reading or viewing material. Also, if we start censoring materials, where does it end? What if someone finds a classic book, such as To Kill a Mockingbird, offensive? Should it be taken away? This would be a huge loss for everyone.</AI\_End> All parents should pick their children movies,music,and magazines they look at and watch, and listen to.

\vspace{1em}
\textbf{Predicted Spans:}

\textcolor{green!60!black}{I don't that the libraries should take awy the books, movies, or any other materials. What if their little kids want to watch a movie or something and they can't because all of the materials are gone.}
\textcolor{red}{I think that the librarians should keep all the materials. If they take away one book, then they have to take away all the other books that someone else might find offensive. This would leave nothing for anyone to read. If someone finds a book, movie, or any other material offensive, they should not take it out. They should not read it or watch it. It's not fair to take away someone else's right to read or watch what they want. Everyone has different opinions and beliefs, and they should be able to express them through their choice of reading or viewing material. Also, if we start censoring materials, where does it end? What if someone finds a classic book, such as To Kill a Mockingbird, offensive? Should it be taken away? This would be a huge loss for everyone.}
\textcolor{green!60!black}{All parents should pick their children movies,music,and magazines they look at and watch, and listen to.}

\vspace{1em}

\begin{minipage}{0.32\textwidth}
    \centering
    \includegraphics[width=\linewidth, height=3.5cm]{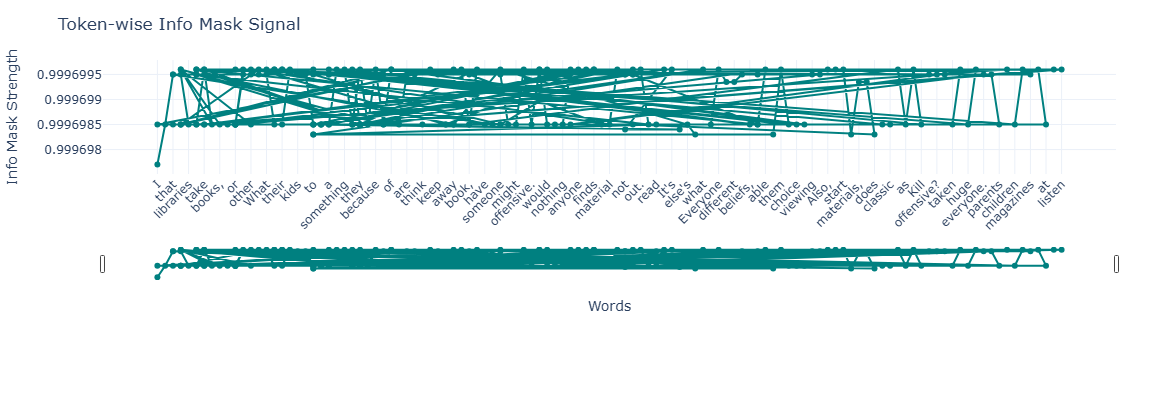}
    \captionof{figure}{Token-wise Info-Mask Signal Strength}
    \label{fig:infomask_strength}
\end{minipage}
\hfill
\begin{minipage}{0.32\textwidth}
    \centering
    \includegraphics[width=\linewidth, height=3.5cm]{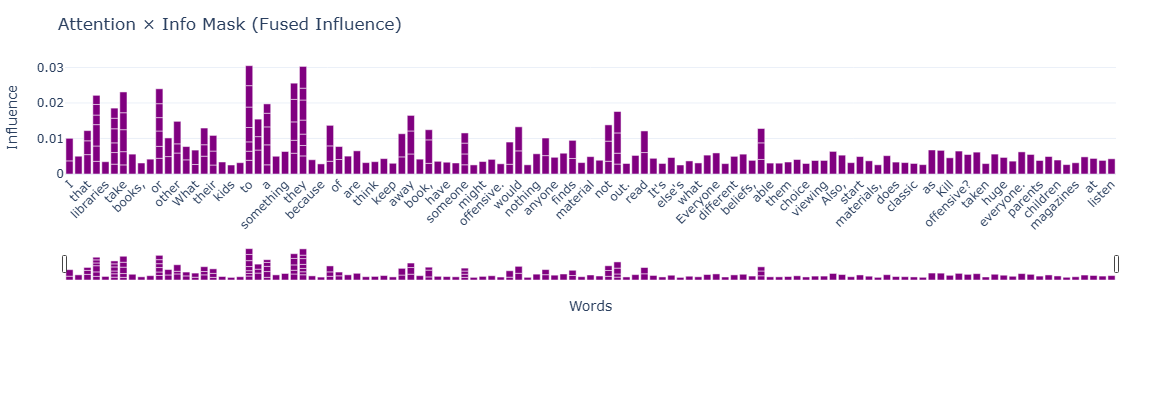}
    \captionof{figure}{Attention × InfoMask Visualization}
    \label{fig:attxinfomask}
\end{minipage}
\hfill
\begin{minipage}{0.32\textwidth}
    \centering
    \includegraphics[width=\linewidth, height=3.5cm]{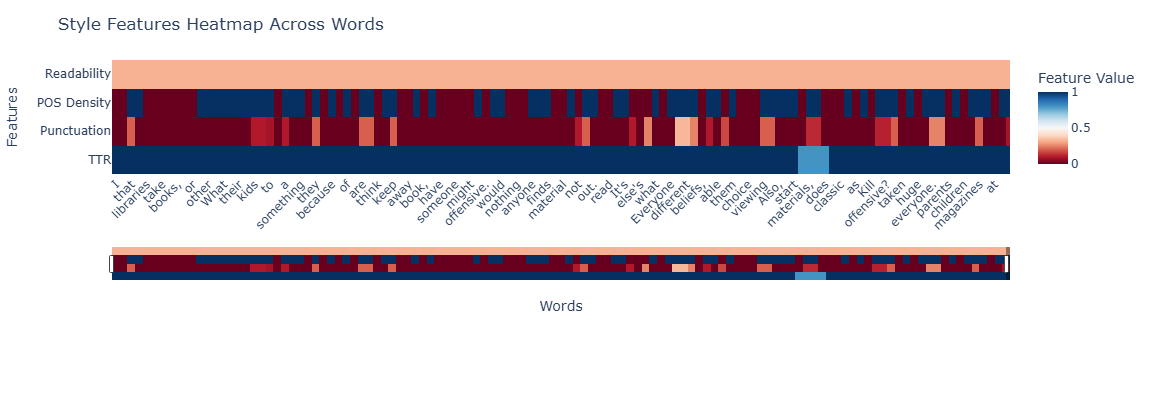}
    \captionof{figure}{Stylometric Features Heatmap}
    \label{fig:styloheatmap}
\end{minipage}

\end{tcolorbox}
\caption{Comparison of Original Text and Model-Predicted Authorship Segmentation. Human-written spans are highlighted in green, while AI-generated spans are marked in red. All corresponding interpretability visualizations are grouped below. (Additional examples will be provided if needed)}
\label{fig:combined-span-analysis}
\end{figure*}

\end{document}